%% file: main.tex
\pdfoutput=1

\documentclass[10pt,twocolumn,letterpaper]{article}

\usepackage[pagenumbers]{cvpr}

\input{preamble}

\definecolor{cvprblue}{rgb}{0.21,0.49,0.74}
\usepackage[pagebackref,breaklinks,colorlinks,allcolors=cvprblue]{hyperref}

\input{linebreak_layout}

\def\confName{CVPR}
\def\confYear{2027}

\title{RePT: 3D Geometry Emerges from 2D Relational Pretraining}

\author{
Xize Duan \and
Binke Hou \and
Xingqiao Huang \and
Xinyuan Guan \and
Jiaxin Huang \and
Kaiyuan Hu \and
Zipeng Sun \and
Hong Kang \and
Meng Bi \and
Xue Liu \and
Yili Jin \and
\parbox{0.9\textwidth}{\centering\vspace{-1mm}{\tt\small xize.duan@mail.mcgill.ca}}
}

\begin{document}
\twocolumn[{%
\renewcommand\twocolumn[1][]{#1}%
\maketitle
\vspace{-10mm}
\begin{center}
\includegraphics[width=\textwidth]{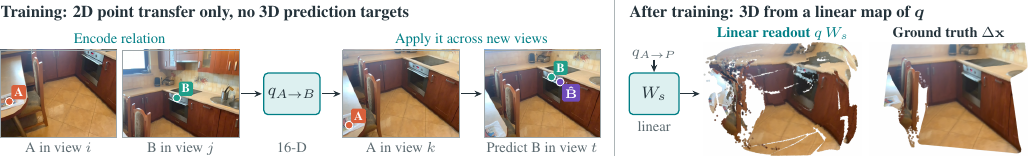}
\captionof{figure}{RePT learns relations between distinct physical points using only cross-view 2D point prediction, with no 3D prediction targets. A linear readout fitted after training maps the 16-D relations to 3D displacements. Point markers are illustrative.}
\label{fig:teaser}
\end{center}
\vspace{-1mm}
}]
\input{0_abstract}
\input{main_content}
{
    \small
    \bibliographystyle{ieeenat_fullname}
    \bibliography{main}
}

\clearpage
\appendix
\input{appendix}

\end{document}

%% file: preamble.tex
\usepackage[T1]{fontenc}
\usepackage{siunitx}
\input{math_commands}


\makeatletter
\IfFormatAtLeastTF{2025-06-01}{%
  \IfFormatAtLeastTF{2026-06-01}{}{%
    \IfPackageAtLeastTF{lineno}{2026/02/11}{%
      \IfPackageAtLeastTF{lineno}{2026/02/17}{}{%
        \let\RePT@originalLNmakecol\@LN@makecol
        \AtBeginDocument{%
          \RemoveFromHook{build/column/before}[lineno]%
          \let\@LN@makecol\RePT@originalLNmakecol
        }%
      }%
    }{}%
  }%
}{}%
\makeatother

%% file: math_commands.tex
\usepackage{amsmath,amsfonts,bm}

\def\eqref#1{equation~\ref{#1}}

\def\1{\bm{1}}

\DeclareMathAlphabet{\mathsfit}{\encodingdefault}{\sfdefault}{m}{sl}
\SetMathAlphabet{\mathsfit}{bold}{\encodingdefault}{\sfdefault}{bx}{n}

%% file: linebreak_layout.tex
\makeatletter
\patchcmd{\BR@backref}{\newblock}{\unskip\nobreak\hspace{0.35em}}{}{%
    \PackageError{RePT}{Could not protect bibliography back-references}{}%
}
\makeatother

\apptocmd{\thebibliography}{%
    \parfillskip=0pt plus 0.30\linewidth\relax
    \emergencystretch=3em\relax
    \tolerance=1000\relax
    \finalhyphendemerits=1000000\relax
}{}{\PackageError{RePT}{Could not refine bibliography line breaking}{}}

%% file: 0_abstract.tex
\begin{abstract}
Learning spatial representations often relies on explicitly predicting geometric quantities,
but how physical geometry is organized within learned representations remains less understood.
We ask whether 3D structure can emerge from learning relations across 2D observations,
without 3D prediction targets.
We introduce RePT, which learns compact relations between distinct physical points through cross-view 2D point prediction.
The learned relation space develops a geometric organization in which magnitudes, angles, and neighborhoods reflect physical 3D displacement.
A single linear readout fitted on 16 scenes recovers displacement length and angle on 39 held-out scenes without refitting (Pearson $r$ = 0.72 and 0.78).
The ARKitScenes-trained checkpoint also retains strong geometric organization on ScanNet++, showcasing generalizability.
In controlled comparisons, distinct-point relational prediction consistently yields stronger geometric organization than matched same-point correspondence learning.
This organization depends on additive relations and adaptive scene context, and higher point-transfer accuracy does not by itself produce stronger organization.
More broadly, our results suggest that 3D geometry need not be an explicit prediction target and can instead emerge as a structural property of representations trained to relate 2D observations, pointing to relational prediction as a complementary route to learning the representations that spatial intelligence builds on.
\end{abstract}

%% file: main_content.tex
\section{Introduction}
Spatial intelligence requires representations that make the structure of the physical world accessible for reasoning and interaction. Existing approaches often obtain explicit geometry by predicting quantities such as depth, camera pose, point maps, or reconstructed scenes for downstream applications~\citep{ranftl2022midas,smith2025flowmap,wang2024dust3r,wang2025vggt,zhang2026d4rt}. At the same time, latent representations learned from purely visual objectives, including reconstruction, correspondence, and cross-view prediction, can also encode substantial geometric information~\citep{weinzaepfel2022croco,weinzaepfel2023crocov2,sajjadi2023rust,mitchel2026xfactor}. Together, these results suggest that geometric structure can arise under a range of different training objectives. In these formulations, geometry is typically anchored to a single observation, as in depth or surface orientation, or to the transformation between views, as in camera motion and cross-view correspondence~\citep{ranftl2022midas,smith2025flowmap,jabri2020space,sarlin2020superglue}.

However, how geometric information is organized in these representations remains less understood. Recent work has probed learned visual features with readouts of varying capacity, showing that coarse geometric quantities, such as relative depth and viewpoint, can often be recovered with linear or lightweight probes~\citep{ma2024imagenet3d,simeoni2025dinov3}, whereas richer metric structure, such as dense 3D geometry, is typically recovered with stronger nonlinear task-specific decoders~\citep{elbanani2024probing,chen2025feat2gs}, and camera-motion structure becomes accessible through a learned adapter~\citep{chen2026seese3}. This suggests an important distinction between a representation that contains recoverable geometric information and one whose latent space is itself geometrically organized. We ask whether metric 3D relations can emerge in a geometrically organized latent space and admit a linear readout without explicit 3D prediction targets.

To study this question, we use point-to-point relations as a minimal relational primitive. Unlike depth or camera motion, a relation between two physical points is defined directly between scene points rather than with respect to a particular viewpoint. Its physical meaning is tied to their underlying 3D displacement, and it can be supervised entirely through their 2D observations across views. This makes point relations a natural testbed for this question.

We instantiate this idea with 2D Relational Pretraining (RePT), which explicitly represents the directed relation between two distinct physical points while training it only through 2D image-space prediction. Given a set of context images of the same physical scene and observations of two physical points $A$ and $B$ in different views, RePT constructs a scene representation $C$ and a compact 16-dimensional directed relation token $q_{A\rightarrow B}$ representing their relationship. Given an observation of $A$ in one context image, the model uses this relation to recover the 2D location of $B$ in another view where it is visible, with all predictions conditioned on $C$. The relational objective predicts only 2D coordinates, without depth, camera-pose, or 3D-coordinate targets.

As a result, the learned relation space develops a coherent metric 3D organization. Consider three physical points $A$, $B$, and $D$. The angle between $q_{A\rightarrow B}$ and $q_{A\rightarrow D}$ closely tracks the angle between their corresponding physical 3D displacements, and the norm $|q_{A\rightarrow D}|_2$ correlates with the physical distance $|\Delta \mathbf{x}_{A\rightarrow D}|_2$ as well. Moreover, a simple linear readout maps relation tokens to metric 3D displacements, yielding coherent relative point clouds. A shared readout preserves this structure in both readout-fitting scenes and scenes excluded from fitting, and the ARKitScenes-trained model retains it on another dataset.

Controlled experiments further clarify what gives rise to this geometric organization. A matched same-point correspondence baseline also develops 3D structure, but RePT remains stronger and degrades less under shared and held-out readouts, suggesting that distinct-point relational prediction promotes stronger displacement-aligned organization than same-point correspondence. Notably, replacing the additive relation parameterization with an MLP improves point-transfer accuracy but substantially weakens geometric organization, showing that better task performance does not necessarily yield a more geometrically organized representation. This organization also depends strongly on adapting the shared context representation during relational training. Auxiliary reconstruction is not necessary for geometric emergence, but improves optimization and point-transfer performance.

These results reveal a complementary route to geometric representation learning. RePT shows that purely 2D relational supervision can induce a representation whose latent organization aligns with physical 3D space through a simple linear readout. This is attractive for scaling spatial representation learning because image-space relational supervision does not require metric 3D annotations: training on RGB-only CoTracker3 tracks yields comparable geometry.

Our contributions are threefold:
\begin{itemize}
\item We introduce 2D relational pretraining to learn directed relations between distinct scene points through cross-view 2D prediction, without predicting 3D geometry.
\item We show that the learned relation space develops a scene-centric 3D organization: its native structure reflects physical displacement, and a simple linear readout recovers coherent metric 3D geometry.
\item We show that this organization is shared across scenes and transfers to unseen environments, and that distinct-point relational prediction strengthens cross-scene consistency over matched same-point correspondence learning.
\end{itemize}

\begin{figure*}[t]
    \centering
    \includegraphics[width=\linewidth]{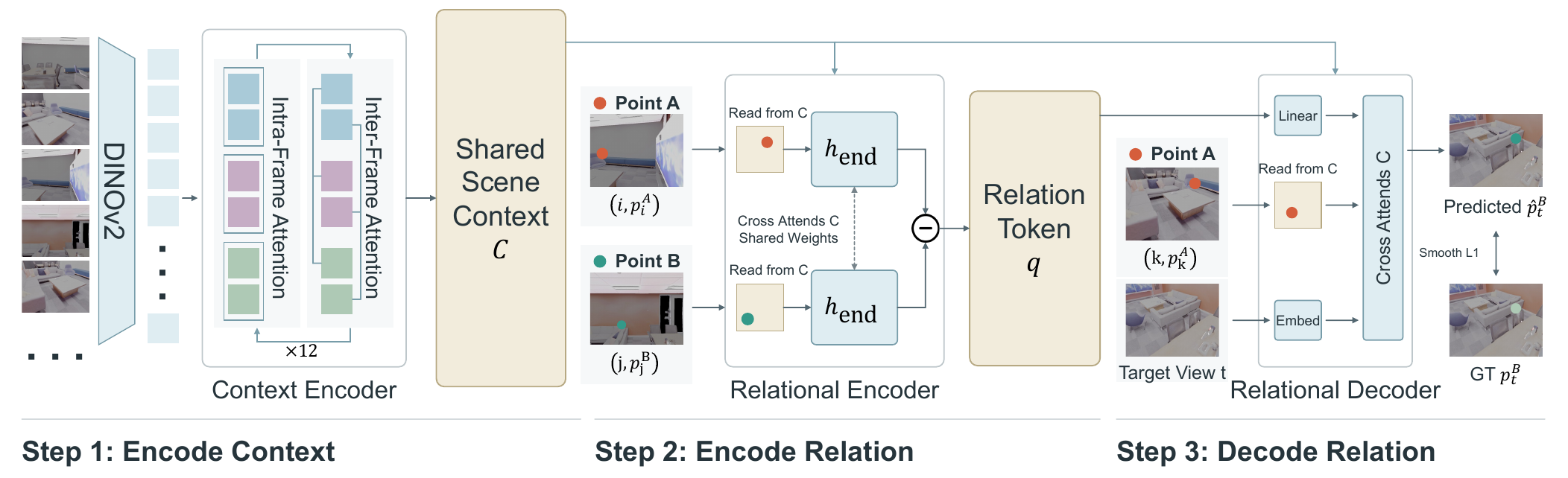}
    \caption{
    Overview of the main task of RePT. RePT constructs a shared scene representation
    $C$, extracts a relation token $q$, transfers the relation to a new pair
    of views, and is trained with a Smooth L1 loss.
    }
    \label{fig:architecture}
\end{figure*}

\section{Related Work}

We review explicit geometric prediction, spatial representations learned from 2D prediction, and geometric structure in learned visual representations.

\subsection{Geometry from Explicit 3D Prediction}
Recent geometric vision models learn spatial representations by directly predicting explicit 3D quantities. DUSt3R~\citep{wang2024dust3r} formulates multi-view reconstruction through dense pointmap prediction, while MASt3R~\citep{leroy2024grounding} extends this formulation with feature matching for improved correspondence and reconstruction. VGGT~\citep{wang2025vggt} further unifies several geometric outputs, jointly predicting camera parameters, depth, point maps, and tracks from multiple images. More recently, D4RT~\citep{zhang2026d4rt} extends this paradigm to dynamic scenes through a unified query-based formulation for recovering 3D point trajectories across space and time. Together, these approaches demonstrate the effectiveness of explicit geometric prediction for learning spatial representations. RePT instead asks what geometric structure can arise when depth, pose, point maps, trajectories, and other 3D quantities are never used as prediction targets.

\subsection{Spatial Representations from 2D Prediction}
A complementary line of work learns spatial structure from relationships between 2D observations. Early context-prediction methods learn visual representations by predicting the relative positions of patches within an image~\citep{doersch2015context}. FlowMap~\citep{smith2025flowmap} recovers depth and camera parameters through optimization against 2D correspondences. CroCo~\citep{weinzaepfel2022croco} and subsequent cross-view masked modeling approaches~\citep{weinzaepfel2023crocov2} reconstruct missing image content with information from another view, encouraging representations to capture geometric dependencies across viewpoints. Correspondence and consistency objectives~\citep{wang2019learning, jabri2020space} encourage observations of the same physical point to remain identifiable across viewpoints, thereby learning geometric consistency through \emph{point identity}. RePT instead learns relations between \emph{distinct} physical points that transfer across views.

Novel-view prediction has also been used to learn scene representations, as in GQN~\citep{eslami2018gqn}, and view-synthesis mappings with minimal 3D inductive bias, as in LVSM~\citep{jin2025lvsm}. Prior work has learned latent scene, pose, or transformation representations through novel-view synthesis and related predictive objectives~\citep{sajjadi2023rust,jiang2025rayzer,wang2026less,mitchel2026xfactor}, with some showing that these latent variables align with physically meaningful camera geometry and transfer across scenes~\citep{mitchel2026xfactor}. RePT instead studies whether 2D prediction induces point-to-point relations whose 3D organization is shared across scenes and accessible through a linear readout.

\subsection{Geometry in Learned Representations}
A growing body of work investigates how geometric structure is encoded in learned visual representations. Linear or lightweight probes have shown that pretrained image and video features expose information about 3D viewpoint, depth, surface geometry, and related  attributes~\citep{ma2024imagenet3d,elbanani2024probing,huang2026much,zhou2026understanding}. Other studies identify geometric regularities directly in feature space, including emergent cross-view correspondence in representations trained for non-geometric objectives~\citep{tang2023emergent} and extreme-view geometry that 3D foundation models acquire without being trained on such views~\citep{zhang2026extremeview}. Together, these results suggest that geometric information can arise without being the explicit prediction target.

More complex readouts reveal richer geometric structure. Feat2GS~\citep{chen2025feat2gs} decodes frozen visual features into 3D Gaussian scenes, while related work studies the recovery of dense scene geometry from pretrained representations~\citep{huang2026much}. SeeSE3~\citep{chen2026seese3} examines relative camera-motion geometry and asks how $\mathrm{SE}(3)$ structure becomes accessible under different feature-space readouts. RePT is complementary to these studies. Rather than asking whether geometric quantities can be recovered by a task-specific deep decoder, we ask whether relational representations themselves become organized in the latent space, and whether the information can be decoded with simple readouts.

\section{2D Relational Pretraining}
We introduce 2D relational pretraining, in which a model learns to transfer relations between distinct scene points through cross-view 2D point prediction.

\subsection{Problem Formulation}
Consider a multi-view scene $\mathcal I=\{I_i\}_{i=1}^{M}$, and let
$A$ and $B$ denote two distinct physical points in the scene.
When visible, their 2D locations in view $i$ are denoted by
$\mathbf p_i^A$ and $\mathbf p_i^B$.
The model observes $A$ at
$\mathbf p_i^A$ in view $i$ and $B$ at $\mathbf p_j^B$ in view $j$.
It is then given another observation of the source point $A$,
$\mathbf p_k^A$ in view $k$, together with a target view $t$, and must predict where the corresponding target point $B$ appears:
\begin{equation}
f\left(
\mathcal I;
(i,\mathbf p_i^A),
(j,\mathbf p_j^B);
(k,\mathbf p_k^A),
t
\right)
\rightarrow
\hat{\mathbf p}_t^B .
\end{equation}
The anchor views $i,j$ and query views $k,t$ may differ. Unlike cross-view correspondence, which transfers the identity of the same physical point across views, relational point transfer requires a relation between two \emph{distinct} physical points to remain applicable under a new view configuration.
The relation token is free to encode whatever information is useful for the point-transfer objective, such as local appearance, neighborhood structure, or other features that facilitate the point matching objective. RePT instantiates this relation as a compact token $q_{A\rightarrow B}$; in \autoref{sec:exp}, we analyze the structure that emerges in this relation space.

\subsection{Architecture}
\label{sec:architecture}

RePT realizes relational point transfer through three core components. The
Context Encoder constructs a shared multi-view scene representation $C$; the
Relational Encoder extracts a compact token $q_{A\rightarrow B}$ from the two
anchor observations; and the Relational Decoder applies this relation to a new
source observation to predict the target point. An auxiliary feature
reconstruction branch is used only during training and is described in
\autoref{sec:training_objectives}.

\paragraph{Context Encoder.}

Firstly, we use DINOv2~\citep{oquab2023dinov2} to map each scene image to patch features
$Z_1,\ldots,Z_M$. The Context Encoder aggregates these features through alternating
within-view and inter-view attention:
\begin{equation}
C = E_{\mathrm{ctx}}(Z_1,\ldots,Z_M).
\end{equation}
The resulting representation $C$ preserves per-view spatial organization while allowing information to be exchanged across views. This provides a shared scene memory from which subsequent modules can read features associated with a
particular view and image location.

\paragraph{Relational Encoder.}

Given an endpoint observation $(m,\mathbf p)$, we first obtain its local scene feature by bilinearly interpolating the feature map of view $m$ in $C$ at location $\mathbf p$:
\begin{equation}
\mathbf c_{m,\mathbf p} = \operatorname{Interp}(C_m,\mathbf p).
\end{equation}
The Relational Encoder concatenates this addressed feature with the point coordinate and view identity, and lets the resulting endpoint token interact with $C$ through four blocks of self-attention followed by cross-attention, to produce an endpoint embedding $h(m,\mathbf p;C)$. For anchor observations $(i,\mathbf p_i^A)$ and $(j,\mathbf p_j^B)$, their directed relation is represented by the difference between the two endpoint embeddings:
\begin{equation}
q_{A\rightarrow B} = h(j,\mathbf p_j^B;C) - h(i,\mathbf p_i^A;C).
\end{equation}
We use a 16-dimensional relation token $q_{A\rightarrow B}$.

\paragraph{Relational Decoder.}

To transfer the learned relation, the Relational Decoder receives the scene
memory $C$, the relation token $q_{A\rightarrow B}$, a new observation
$(k,\mathbf p_k^A)$ of the source point, and the target-view index $t$. It
reads the source-point feature from $C$ via bilinear interpolation, combines it with the relation token and
target-view information, and predicts the location of $B$ via cross attention:
\begin{equation}
\hat{\mathbf p}_t^B
=
D_{\mathrm{rel}}
\left(
C,
q_{A\rightarrow B},
k,
\mathbf p_k^A,
t
\right).
\end{equation}
The same relation token can therefore be applied across different source and target views, with $C$ providing the scene context needed for cross-view transfer.

\subsection{Training Objectives}
\label{sec:training_objectives}

Relational point transfer is the primary objective. Cross-view feature
reconstruction additionally regularizes $C$.

\paragraph{Relational point transfer.}
Given the predicted target coordinate $\hat{\mathbf p}_t^B$ and its
ground-truth location $\mathbf p_t^B$, we optimize
\begin{equation}
\mathcal L_{\mathrm{rel}}
=
\operatorname{SmoothL1}
\left(
\hat{\mathbf p}_t^B,
\mathbf p_t^B
\right).
\end{equation}
We normalize coordinates to \mbox{$168\times224$} inputs.%

\paragraph{Auxiliary feature reconstruction.}
During training, we optionally regularize $C$ with a CroCo-style~\citep{weinzaepfel2022croco} cross-view
feature reconstruction objective using $C$ as condition. For this branch, an extra view is excluded when constructing $C$, and its frozen visual features $Z_{\mathrm{tar}}$ are partially masked. A lightweight decoder attends to $C$ and the visible target features to reconstruct the missing features:
\begin{equation}
\hat Z_{\mathrm{tar}}= D_{\mathrm{rec}} \left( C, \widetilde Z_{\mathrm{tar}} \right).
\end{equation}
The reconstruction loss $\mathcal L_{\mathrm{rec}}$ is evaluated only on masked patches and combines cosine feature similarity with layer-normalized feature regression. We mask $90\%$ of target patches and use complete masking for $25\%$ of training samples. Our controlled experiments in \autoref{sec:geometric_analysis} show that this objective mainly supports optimization efficiency.

The complete training objective is
\begin{equation}
\mathcal L = \mathcal L_{\mathrm{rel}} + \lambda_{\mathrm{rec}}\mathcal L_{\mathrm{rec}},
\qquad \lambda_{\mathrm{rec}}=0.1.
\end{equation}

\section{Experiments}
\label{sec:exp}

In this section, we study what geometric structure emerges in the learned representations and what gives rise to it. We first analyze the 3D organization of the relation tokens $q$, and then use controlled experiments to identify the learning signals responsible for this emergence. Track construction and evaluation details are in Appendices~\ref{app:exp_details} and~\ref{app:linear_readout}.

\subsection{Experimental Setup}
We train and evaluate RePT on ARKitScenes~\citep{baruch2021arkitscenes}, using 4492 sequences for training and 55 for validation. We construct sparse multi-view landmark tracks using the provided depth maps and camera poses, yielding persistent point identities and their 2D observations across views. Depth and pose are used only for offline track construction and are never provided to the model or used as prediction targets. Track construction details are provided in Appendix~\ref{app:track_construction}.
We also fully train a separate RePT model using CoTracker3~\citep{karaev2025cotracker3}
RGB-only tracks; see \mbox{\autoref{tab:rgb_only_geometry}}.

Each training episode uses 12 context images to construct the scene representation $C$. We then sample two distinct landmarks $A$ and $B$ that are visible across multiple context views. During training, four cross-view observation pairs of $(A,B)$ are used to encode the relation which are averaged to form $q_{A\rightarrow B}$. The resulting relation token is applied to up to eight additional observation pairs as point-transfer queries. The relation and query pairs are distinct. For geometric analysis, we evaluate relation tokens encoded from individual observation pairs without averaging.

For efficiency, we initialize the Context Encoder with 5 epochs of cross-view reconstruction pretraining before relational training as a shared starting point unless otherwise specified, and its effect is studied in \autoref{sec:geometric_analysis}.

\subsection{Emergent 3D Geometry in the Relation Space}
\label{sec:emergent_geometry}

We first ask whether 3D geometric structure emerges directly in the learned relation representation, despite the absence of an explicit 3D prediction
target. For each directed relation
$q_{A\rightarrow B}\in\mathbb{R}^{16}$, let
\[
    \Delta\mathbf{x}_{A\rightarrow B}
    = \mathbf{x}_B-\mathbf{x}_A
\]
denote the corresponding physical 3D displacement. We examine both the native
geometry of $q$ and the structure exposed by a simple linear readout.

\begin{figure*}[t]
    \centering
    \captionsetup[subfigure]{font=small, skip=2pt}

    \begin{subfigure}[t]{0.495\linewidth}
        \centering
        \includegraphics[width=\linewidth]{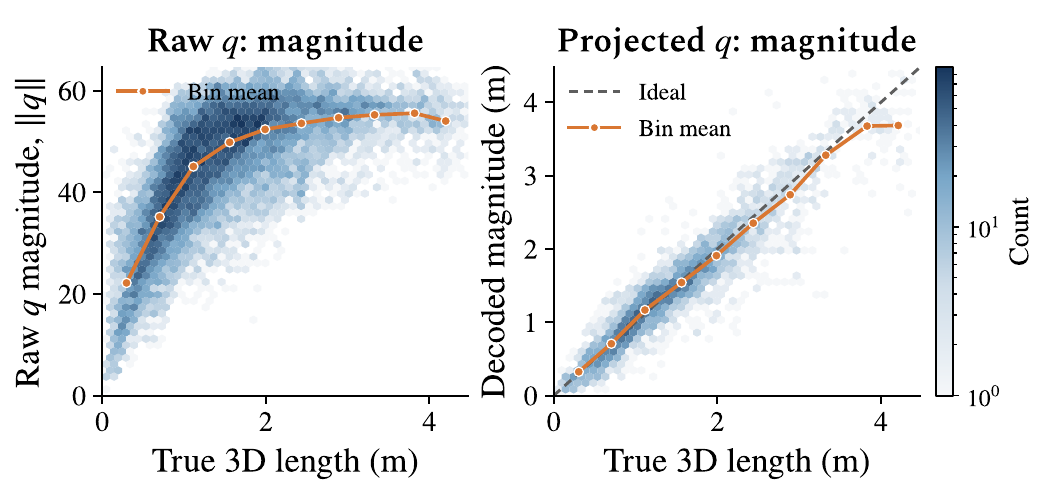}
        \caption{Magnitude}
        \label{fig:q_magnitude}
    \end{subfigure}
    \hfill
    \begin{subfigure}[t]{0.495\linewidth}
        \centering
        \includegraphics[width=\linewidth]{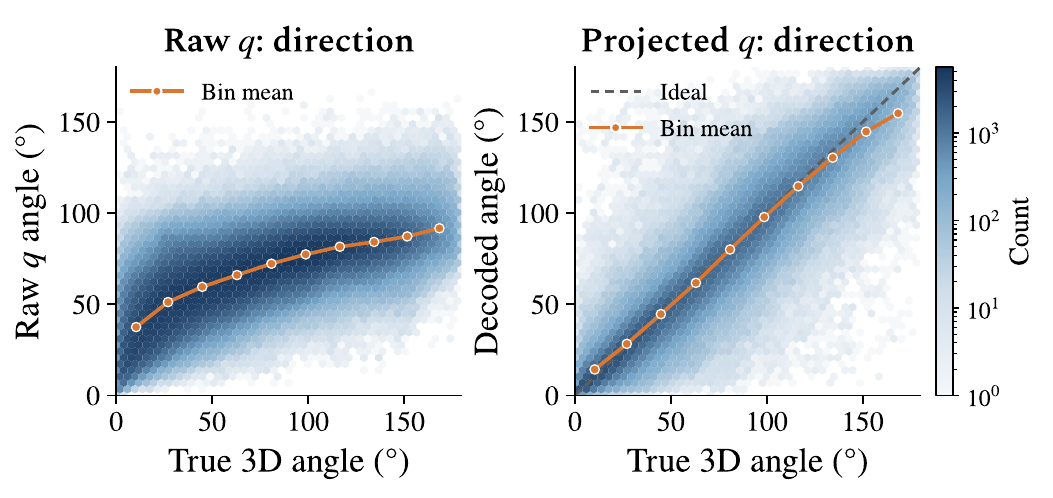}
        \caption{Direction}
        \label{fig:q_direction}
    \end{subfigure}

    \caption{
        \textbf{Geometric structure of the relational representation.}
        Raw $q$ already correlates with 3D geometry, while a linear projection
        recovers metric magnitude and direction.
    }
    \label{fig:q_geometry_correlation}
\end{figure*}

\paragraph{Geometry in the raw relation space.}
The native geometry of $q$ already exhibits a clear correspondence with physical 3D displacement. Relation magnitude increases systematically with the physical distance between the reference and target landmarks, while pairwise angles between relation tokens track the corresponding angles between 3D displacement vectors, as shown in \autoref{fig:q_geometry_correlation} and \autoref{tab:relation_geometry}. Raw $q$ achieves a length Pearson correlation of 0.5921 and an angle Pearson correlation of 0.6789. Local neighborhood is also partially preserved with an NN-Overlap@10 of 0.2639.

In this setting, latent magnitudes and directions are distorted relative to physical 3D space and saturate at a certain level. Nevertheless, the presence of both metric and directional correlations and neighborhood preservation in the native 16-dimensional representation shows that relational prediction already induces substantial 3D organization before any explicit geometric calibration is applied.

\begin{table*}[t]
      \centering
      \caption{
          \textbf{Geometric organization of the RePT relation space under increasingly restrictive readouts.}
          Results are averaged over 300 random 16/39 scene splits.
          Raw relation tokens already reflect physical 3D displacement.
          Scene-specific linear calibration recovers strong metric geometry,
          while a shared linear readout retains substantial structure on the readout-fitting scenes and transfers to the held-out scenes. Shared and transfer evaluation use the same frozen $W$.
      }
      \label{tab:relation_geometry}
      \vspace{-1mm}

      \scriptsize
      \renewcommand{\arraystretch}{0.92}
      \setlength{\tabcolsep}{4pt}

      \begin{tabular*}{\linewidth}{
          @{\extracolsep{\fill}}
          lcccccc
          @{}
      }
          \toprule
          &
          \multicolumn{2}{c}{Length}
          & \multicolumn{2}{c}{Angle}
          & NN-Overlap@10 $\uparrow$
          & 3D Error (m) $\downarrow$
          \\
          \cmidrule(lr){2-3}
          \cmidrule(lr){4-5}

          Representation / Readout
          & Pearson $\uparrow$
          & Spearman $\uparrow$
          & Pearson $\uparrow$
          & Spearman $\uparrow$
          & &
          \\
          \midrule

          Raw $q$
          & 0.5921
          & 0.6168
          & 0.6789
          & 0.6822
          & 0.2639
          & -- \\

          \midrule

          Scene-specific $W_s$
          & 0.8639
          & 0.8779
          & 0.8804
          & 0.8818
          & 0.4657
          & 0.5136 \\

          Shared $W$ (16 fit) + scene rotation
          & 0.7349
          & 0.7523
          & 0.7979
          & 0.8001
          & 0.3721
          & 0.8637 \\

          Transfer $W$ (39 held-out) + scene rotation
          & 0.7168
          & 0.7349
          & 0.7844
          & 0.7867
          & 0.3604
          & 0.8962 \\

          \bottomrule
      \end{tabular*}
      \vspace{-1mm}
  \end{table*}

\paragraph{Linearly accessible 3D geometry.}
We next ask whether the remaining distortion can be calibrated by a simple linear transformation. For each scene, we fit a linear map
$W_s:\mathbb{R}^{16}\rightarrow\mathbb{R}^{3}$ directly from relation tokens to their corresponding physical 3D displacement vectors:
\[
    q_{A\rightarrow P} W_s 
    \approx
    \Delta\mathbf{x}_{A\rightarrow P}.
\]
For each scene, we randomly sample 1,500 valid cross-view relation pairs $q_{A\rightarrow B}$ without fixing an anchor point. We fit $W_s$ on 80\% of the relations and test on the remaining 20\%.

The linear projection substantially strengthens geometric agreement. In \autoref{tab:relation_geometry}, length Pearson correlation increases from 0.5921 to 0.8639, while angle Pearson correlation increases from 0.6789 to 0.8804. Neighborhood preservation also improves from 0.2639 to 0.4657, and the predicted displacement vectors achieve a mean 3D error of 0.5136\,m.

\begin{figure*}[t]
    \centering
    \captionsetup[subfigure]{font=small, skip=2pt}

    \begin{subfigure}[t]{0.49\linewidth}
        \centering
        \setlength{\tabcolsep}{1pt}
        \begin{tabular}{ccc}
            \scriptsize Ref. &
            \scriptsize GT &
            \scriptsize Latent \\[-1pt]
            \includegraphics[width=0.315\linewidth]{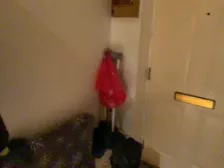} &
            \includegraphics[width=0.315\linewidth]{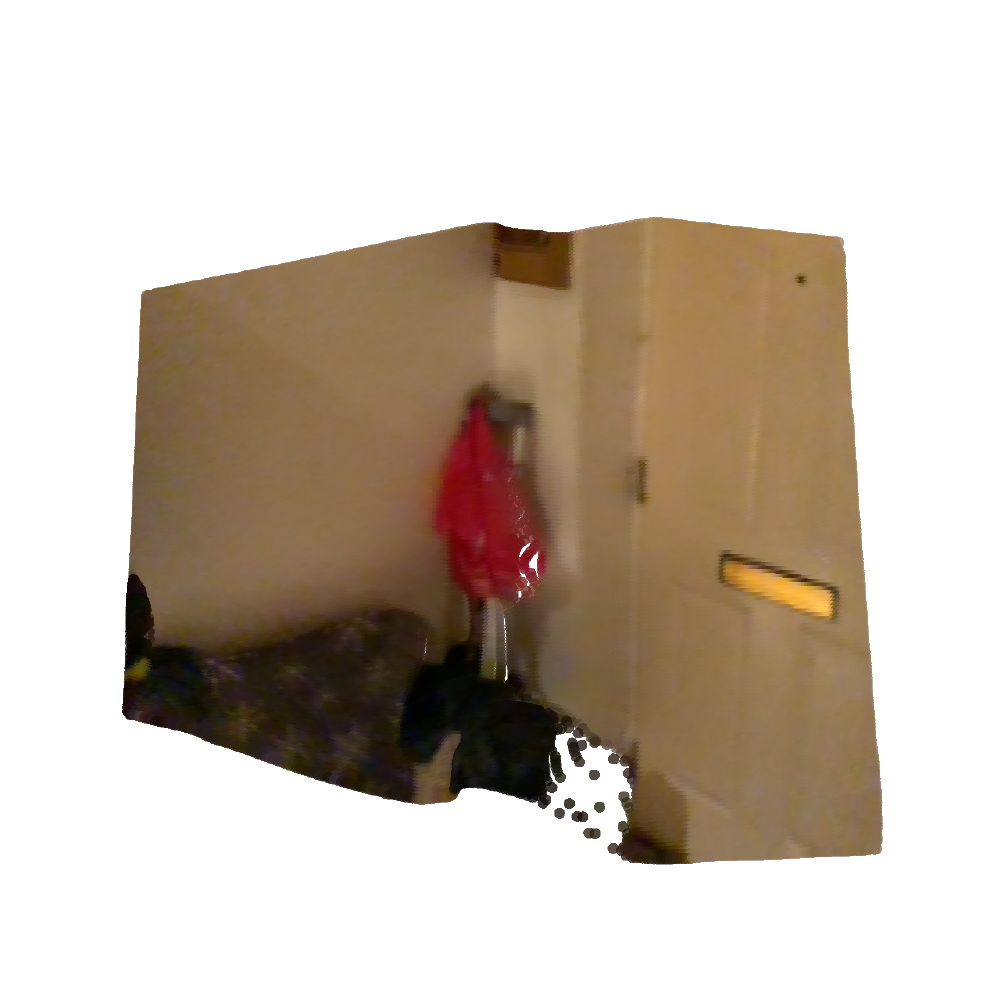} &
            \includegraphics[width=0.315\linewidth]{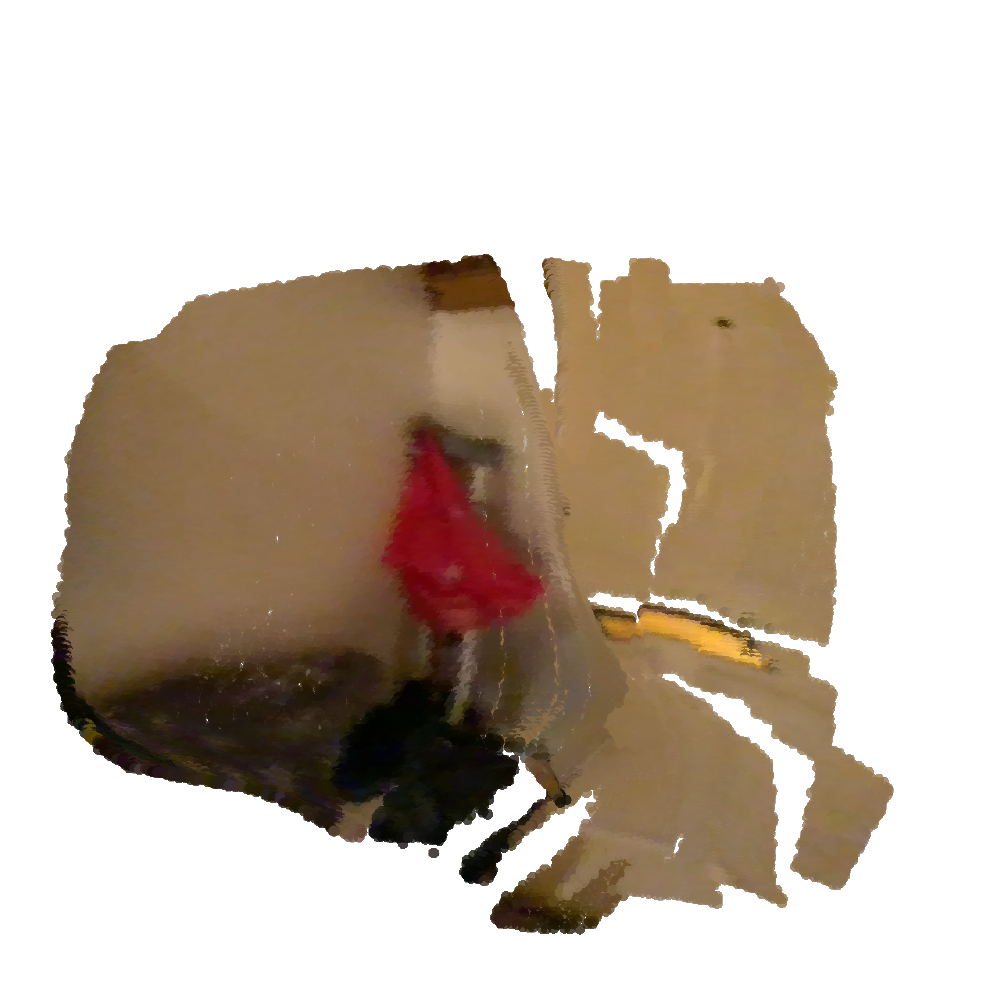}
        \end{tabular}
        \caption{}
    \end{subfigure}
    \hfill
    \vspace{-3pt}
    \begin{subfigure}[t]{0.49\linewidth}
        \centering
        \setlength{\tabcolsep}{1pt}
        \begin{tabular}{ccc}
            \scriptsize Ref. &
            \scriptsize GT &
            \scriptsize Latent \\[-1pt]
            \includegraphics[width=0.315\linewidth]{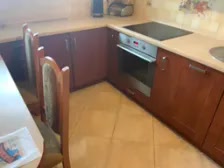} &
            \includegraphics[width=0.315\linewidth]{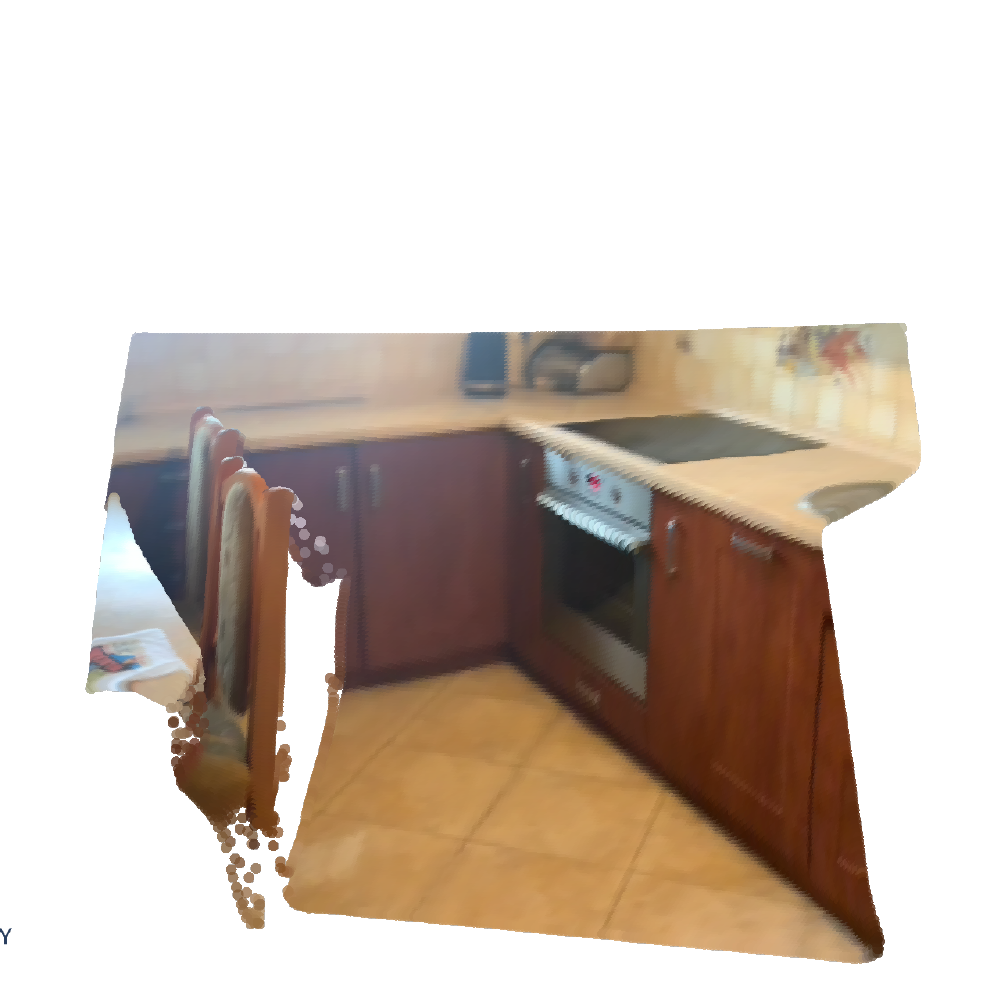} &
            \includegraphics[width=0.315\linewidth]{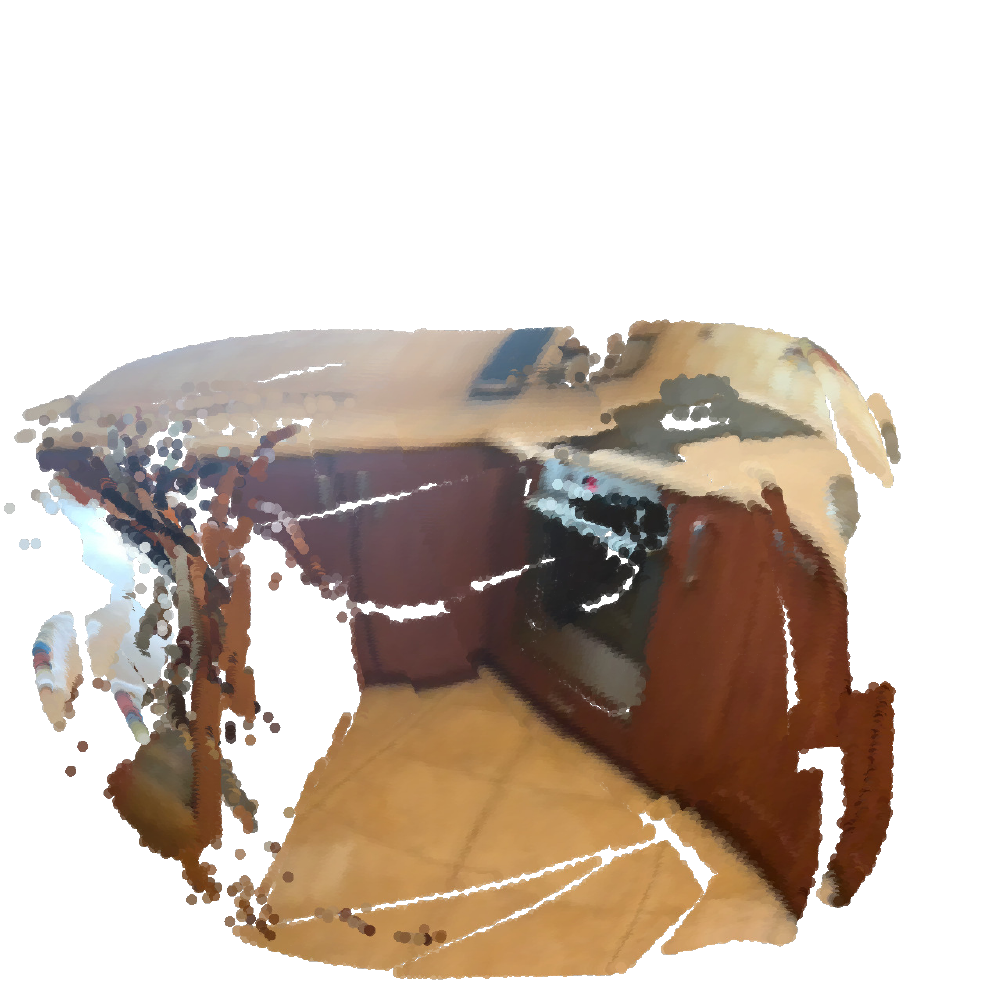}
        \end{tabular}
        \caption{}
    \end{subfigure}

    \begin{subfigure}[t]{0.49\linewidth}
        \centering
        \setlength{\tabcolsep}{1pt}
        \begin{tabular}{ccc}
            \includegraphics[width=0.315\linewidth]{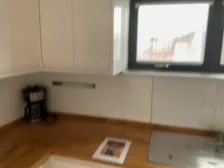} &
            \includegraphics[width=0.315\linewidth]{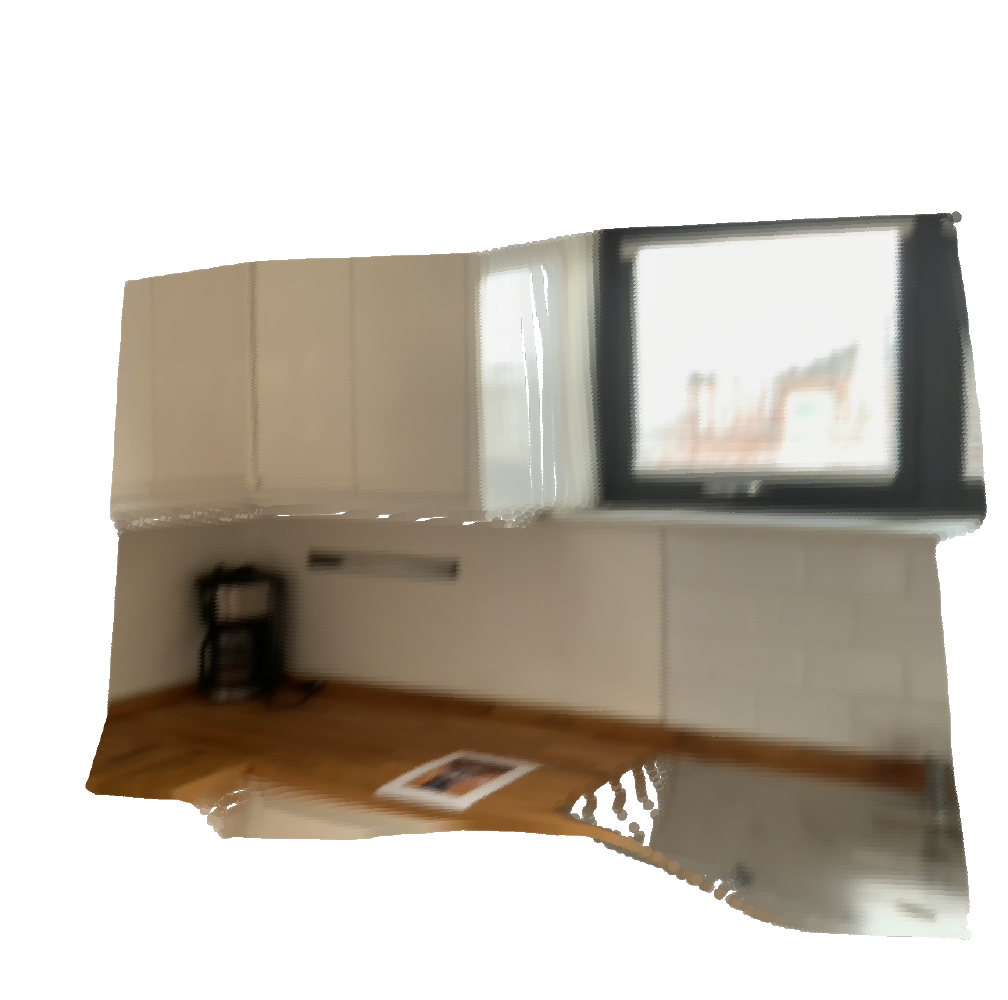} &
            \includegraphics[width=0.315\linewidth]{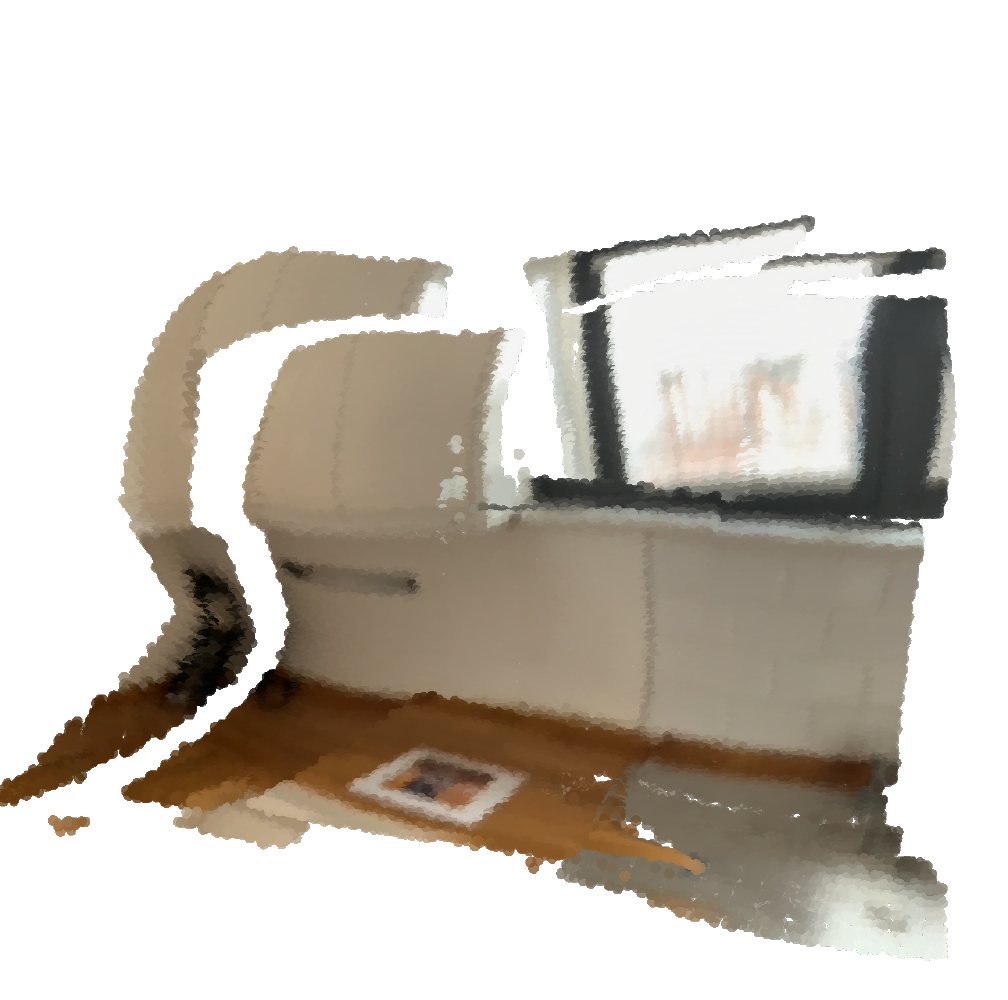}
        \end{tabular}
        \caption{}
    \end{subfigure}
    \hfill
    \begin{subfigure}[t]{0.49\linewidth}
        \centering
        \setlength{\tabcolsep}{1pt}
        \begin{tabular}{ccc}
            \includegraphics[width=0.315\linewidth, angle=180, origin=c]{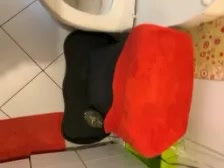} &
            \includegraphics[width=0.315\linewidth, angle=90, origin=c]{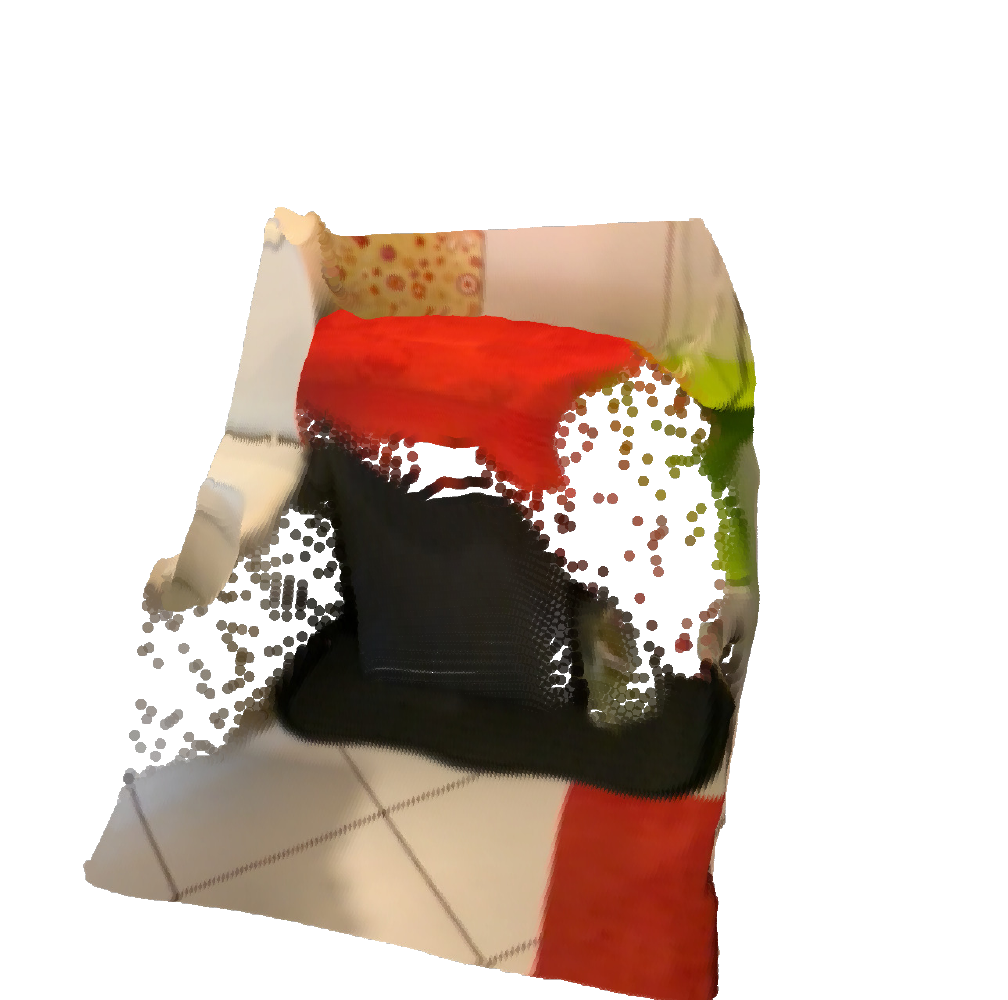} &
            \includegraphics[width=0.315\linewidth, angle=90, origin=c]{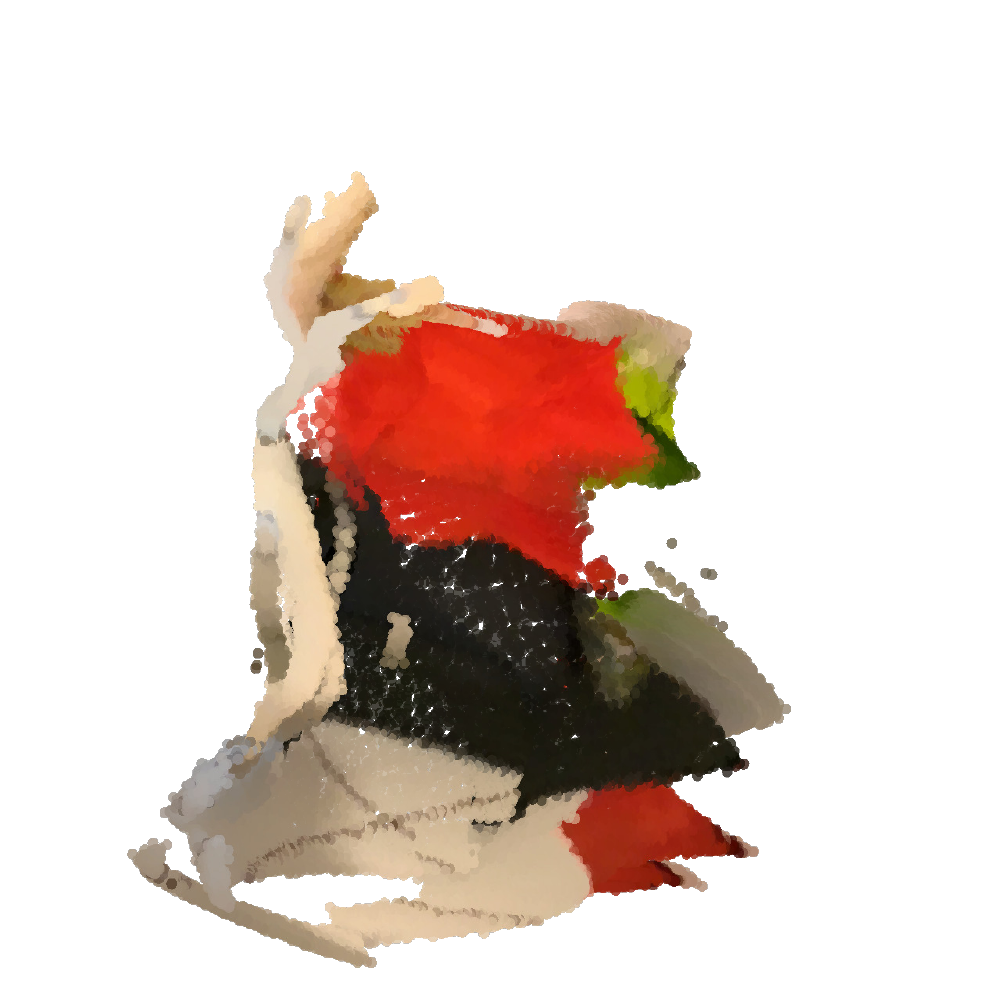}
        \end{tabular}
        \caption{}
    \end{subfigure}

    \caption{
        \textbf{Emergent geometric structure in the relational latent space.}
        Each example shows a reference image, the ground-truth 3D geometry,
        and the corresponding geometry obtained by projecting the learned
        relational representation linearly into 3D.
    }
    \label{fig:latent_geometry_vis}
\end{figure*}

The linearly decoded geometry is also visible qualitatively. For each example, we choose a reference point $A$ and visualize the predicted displacements $q_{A\rightarrow P} W_s$ to a set of target points $P$, placing $A$ at the origin. The decoded displacements form a relative point cloud around the reference point. As shown in \autoref{fig:latent_geometry_vis}, these point clouds recover the dominant structure of the corresponding ground-truth 3D layouts, providing a qualitative view of the geometry exposed by the linear readout.

Overall, these results show that RePT develops substantial geometric structure directly in its raw relation space, while a simple linear readout exposes a strong metric and directional 3D organization within individual scenes.

\subsection{Shared Geometry Across Scenes}
\label{sec:shared_geometry}

We test whether a single linear map recovers geometry across multiple
scenes, all excluded from RePT training.

\paragraph{A shared linear readout across scenes.}
We first ask whether relation spaces from different scenes admit a common
linear mapping to 3D. Because ARKitScenes defines the world coordinate frame
of each sequence independently, relative 3D displacements are comparable
across scenes only up to a scene-specific rotation. We therefore learn a
bias-free shared linear map
$W:\mathbb{R}^{16}\rightarrow\mathbb{R}^{3}$ together with a rotational gauge
$R_s\in\mathrm{SO}(3)$ for each scene by minimizing:
\[
\min_{W,\{R_s\}}
\sum_s
\left\|
Q_s W R_s^\top-D_s
\right\|_F^2 .
\]
Translation does not enter this alignment because $D_s$ contains only relative displacements.

We evaluate this readout over 300 random partitions of the 55 evaluation
scenes into 16 readout-fitting scenes and 39 held-out scenes. We sample 1,500 relation pairs from each fitting
scene, use 80\% to jointly estimate $W$ and
$R_s$, and evaluate the fitted parameters on the remaining 20\%.
Despite using a shared $W$, RePT retains substantial geometric structure: \autoref{tab:relation_geometry} reports length and angle correlations of 0.7349 and 0.7979, respectively, with a 3D endpoint error of 0.8637\,m. This indicates that the relation spaces learned in different scenes admit a common linear 3D readout up to their scene-coordinate rotational gauges.

\paragraph{Transfer to held-out scenes.}
We next test whether the same readout transfers to scenes that did not
participate in fitting $W$. For each 16/39 partition, we freeze the shared
$W$ learned from the 16 fitting scenes and apply it unchanged to all 39
held-out scenes. Length and angle are rotation-invariant and can therefore be
evaluated directly. For 3D endpoint error, which depends on the scene
coordinate frame, we estimate only the scene-specific rotation $R_s$ from
80\% of the relations in each held-out scene and evaluate on the remaining
20\%. $W$ remains fixed throughout.

The last two rows of \autoref{tab:relation_geometry} show that performance
changes only modestly under this transfer setting. Length
correlation decreases from 0.7349 to 0.7168, angle correlation from 0.7979 to
0.7844, and 3D endpoint error increases from 0.8637\,m to 0.8962\,m. Thus, a linear readout estimated from only
16 scenes transfers with little degradation to the remaining 39 scenes,
providing evidence that RePT develops a consistent cross-scene geometric
organization rather than one specific to the fitting scenes.

\paragraph{Cross-dataset generalization.}
We evaluate the same ARKitScenes-trained RePT checkpoint on
ScanNet++~\citep{yeshwanth2023scannetpp}, using the same preprocessing and
geometric evaluation protocol as on ARKitScenes.
We sample four subsets of 55 scenes from the same 537-scene candidate pool
using fixed random seeds. Each subset is evaluated over 300 random 16/39
scene splits, and we report results averaged over the four subsets.
As summarized in \autoref{tab:cross_dataset_geometry}, RePT retains strong geometric organization, with scene-specific length and angle correlations
of 0.8503 and 0.8593.
Under the shared readout, these correlations reach 0.7166 and 0.7637 and
decrease only modestly to 0.6973 and 0.7473 on held-out scenes.
These results show that the geometric organization learned by RePT
persists under dataset shift.

\begin{table}[t]
    \centering
    \caption{
        \textbf{Cross-dataset and RGB-only geometric organization.}
        Scene-averaged Pearson correlations. All models are trained on ARKitScenes; tracks are training correspondences.
    }
    \label{tab:cross_dataset_geometry}
    \label{tab:rgb_only_geometry}
    \vspace{-1mm}
    \scriptsize
    \setlength{\tabcolsep}{2pt}
    \begin{tabular*}{\columnwidth}{@{\extracolsep{\fill}}lcccccc@{}}
        \toprule
        & \multicolumn{4}{c}{Projection tracks}
        & \multicolumn{2}{c}{CoTracker3 tracks} \\
        \cmidrule(lr){2-5}\cmidrule(lr){6-7}
        & \multicolumn{2}{c}{ARKitScenes}
        & \multicolumn{2}{c}{ScanNet++}
        & \multicolumn{2}{c}{ARKitScenes} \\
        \cmidrule(lr){2-3}\cmidrule(lr){4-5}\cmidrule(lr){6-7}
        Readout & Len. & Ang. & Len. & Ang. & Len. & Ang. \\
        \midrule
        Raw $q$ & 0.5921 & 0.6789 & 0.5708 & 0.6453 & 0.6276 & 0.7088 \\
        Scene-specific $W_s$ & 0.8639 & 0.8804 & 0.8503 & 0.8593 & 0.8598 & 0.8558 \\
        Shared $W$ & 0.7349 & 0.7979 & 0.7166 & 0.7637 & 0.7286 & 0.7816 \\
        Transfer $W$ & 0.7168 & 0.7844 & 0.6973 & 0.7473 & 0.7093 & 0.7683 \\
        \bottomrule
    \end{tabular*}
    \vspace{-1mm}
\end{table}

\begin{table*}[t]
    \centering
    \caption{
        \textbf{Controlled comparison of design factors.}
        Results are averaged over 300 random 16/39 scene splits.
        Len.\ and Ang.\ denote scene-averaged Pearson correlations.
        3D Err.\ is the mean endpoint error after scene-rotation calibration, in meters.
        The same fitted shared $W$ is used for both the fit and held-out scenes.
    }
    \label{tab:controlled_ablation}
    \vspace{-1mm}

    \scriptsize
    \renewcommand{\arraystretch}{0.94}
    \setlength{\tabcolsep}{2.15pt}

    \begin{tabular*}{\textwidth}{
        @{\extracolsep{\fill}}
        l
        ccc
        ccc
        ccc
        c
        @{}
    }
        \toprule
        & \multicolumn{3}{c}{Scene-specific $W_s$}
        & \multicolumn{3}{c}{Shared $W$ (16 fit)}
        & \multicolumn{3}{c}{Transfer $W$ (39 held-out)}
        & PCK@4 $\uparrow$ \\
        \cmidrule(lr){2-4}
        \cmidrule(lr){5-7}
        \cmidrule(lr){8-10}

        Method
        & Len. $\uparrow$
        & Ang. $\uparrow$
        & 3D Err. $\downarrow$
        & Len. $\uparrow$
        & Ang. $\uparrow$
        & 3D Err. $\downarrow$
        & Len. $\uparrow$
        & Ang. $\uparrow$
        & 3D Err. $\downarrow$
        & \\
        \midrule

        Full RePT
        & \textbf{0.8639}
        & \textbf{0.8804}
        & \textbf{0.5136}
        & \textbf{0.7349}
        & \textbf{0.7979}
        & \textbf{0.8637}
        & \textbf{0.7168}
        & \textbf{0.7844}
        & \textbf{0.8962}
        & 0.2789 \\

        Correspondence
        & 0.7979
        & 0.8259
        & 0.6383
        & 0.6533
        & 0.7137
        & 0.9670
        & 0.6038
        & 0.6809
        & 1.0150
        & 0.2034 \\

        DINOv2 as $C$
        & 0.6516
        & 0.6616
        & 0.8928
        & 0.4396
        & 0.4728
        & 1.2781
        & 0.4076
        & 0.4434
        & 1.3249
        & 0.0567 \\

        Frozen context
        & 0.2523
        & 0.3175
        & 1.4336
        & 0.1357
        & 0.2268
        & 1.5799
        & 0.1168
        & 0.2088
        & 1.5990
        & 0.0667 \\

        MLP relation
        & 0.3335
        & 0.4051
        & 1.2928
        & 0.2666
        & 0.3781
        & 1.3808
        & 0.2577
        & 0.3729
        & 1.3897
        & \textbf{0.3083} \\

        \addlinespace
        \multicolumn{11}{@{}l}{\textit{Training from scratch}} \\

        Relation + recon.
        & 0.8659
        & \textbf{0.8820}
        & 0.5109
        & 0.7341
        & \textbf{0.7991}
        & 0.8642
        & 0.7159
        & \textbf{0.7857}
        & 0.8964
        & \textbf{0.3055} \\

        Relation only
        & \textbf{0.8673}
        & 0.8808
        & \textbf{0.5102}
        & \textbf{0.7363}
        & 0.7989
        & \textbf{0.8601}
        & \textbf{0.7181}
        & 0.7851
        & \textbf{0.8930}
        & 0.2742 \\

        \bottomrule
    \end{tabular*}

    \vspace{-1mm}
\end{table*}

\subsection{What Promotes Geometric Organization?}
\label{sec:geometric_analysis}

We next ask which aspects of RePT give rise to the geometric organization
observed above. We compare variants that alter the learning objective, the
scene representation, the relation parameterization, and the auxiliary
reconstruction objective. These comparisons, reported in \mbox{\autoref{tab:controlled_ablation}}, distinguish
geometric organization from point-transfer accuracy.

\paragraph{Relational prediction.}
We first compare RePT with a matched same-point correspondence baseline.
Instead of predicting a relation between two distinct landmarks, this model
encodes a single landmark with the same endpoint encoder and directly uses its
embedding to localize the same physical point in another view. For geometric evaluation, we form pairwise differences of the learned
endpoint embeddings and use the same linear readouts as RePT.

The correspondence baseline develops geometric structure under a scene-specific readout, but achieves lower length and angle correlations than RePT. This gap widens under shared and held-out transfer readouts, suggesting that learning relations between distinct physical points promotes more consistent geometric organization across scenes.

\paragraph{Relation parameterization.}
We also test whether the additive relation parameterization contributes to the
emergent geometry. Replacing
$q_{A\rightarrow B}=h(B;C)-h(A;C)$ with a learned MLP relation encoder causes
a large reduction in geometric organization. Scene-specific length and angle correlations decrease, and the gap persists under shared and held-out transfer readouts.

Notably, this variant achieves a higher PCK@4 than the full RePT model. This shows that strong task performance does not by itself imply a geometrically organized latent space. The additive difference structure therefore acts as an inductive bias that favors a more displacement-like organization of the relation space.

\paragraph{Relation dimensionality.}
Across relation dimensions of 3, 4, 8, 16, and 32, RePT consistently develops geometric structure. Increasing the dimension strengthens scene-specific length correlation, while the strongest shared and held-out correlations occur at 3 dimensions. We default to 16 because it balances scene-specific and cross-scene organization; see Appendix~\ref{app:relation_dimensionality}.

\paragraph{Context adaptation.}
We next examine the role of the shared scene representation $C$. Replacing the
learned Context Encoder with frozen DINOv2~\citep{oquab2023dinov2} features substantially weakens
geometric organization across all three readout settings. Although this variant
retains some scene-specific structure, its length and angle correlations fall
from 0.8639 and 0.8804 to 0.6516 and 0.6616, with larger degradation under
shared and held-out transfer evaluation.

The effect is stronger when we initialize the Context Encoder with
reconstruction pretraining but freeze it during relational training. This
variant preserves the architecture and initialization of RePT, yet its
scene-specific correlations fall to 0.2523 and 0.3175, with similarly weak
shared and transferred readouts. Thus, a fixed reconstruction-pretrained scene representation is insufficient to produce the geometric organization observed in full RePT, highlighting the importance of jointly adapting $C$ to the relational objective. Additional probes in Appendix~\ref{app:c_geometry} show that $C$ supports more accurate depth, camera-pose, and spatial-relation prediction than frozen DINOv2 features.

\paragraph{Reconstruction.}
To isolate the reconstruction objective, we train matched models from scratch with and without reconstruction, and both develop comparable scene-specific and shared geometry. Reconstruction mainly improves PCK and accelerates optimization. We therefore use it to facilitate training and target localization.

\paragraph{RGB-only correspondence construction.}
We fully train RePT using CoTracker3 correspondences on randomly sampled RGB
pixels, keeping the remaining pipeline unchanged. \autoref{tab:rgb_only_geometry} shows comparable geometry across readout
settings, indicating that geometric organization
also emerges with RGB-only tracking correspondences. Projection-based
construction remains our default because preprocessing is substantially faster.
Appendix~\ref{app:rgb_only_tracks} provides details and metrics.

Overall, distinct-point relational prediction and additive relation parameterization promote the geometric organization observed in RePT. This organization also depends strongly on adapting the shared scene representation during relational training. Reconstruction aids optimization but is not needed for \mbox{geometric emergence}.

\section{Discussion and Limitations}

Geometry need not be supervised to be learned. RePT never predicts depth, pose, or 3D points, yet its relation space becomes organized by metric 3D displacement. This organization is not an artifact of the readout. Under the same scene-specific protocol, frozen-context and MLP-relation variants reach length and angle correlations of only 0.25 and 0.32, and 0.33 and 0.41, compared with 0.86 and 0.88 for RePT. The MLP-relation variant is the clearest case, attaining the highest PCK@4 of all variants (0.31 vs.\ 0.28) while losing most of the geometry. Solving relational transfer and organizing the latent space geometrically are therefore distinct outcomes, and which one occurs depends on how relations are learned, not only on the task.

RePT also takes a different view of what spatial understanding requires. Reconstruction methods treat it as recovering coordinates for every pixel or point, which calls for dense 3D supervision or per-scene optimization and produces far more geometry than most spatial queries need. Human spatial perception suggests that this is not a prerequisite. People act effectively in space without a globally consistent metric map, relying instead on spatial knowledge that is built up through continued observation and organized around relations between places~\citep{warren2017wormholes}. RePT follows the same principle. It learns from 2D point observations, which RGB-only tracking can supply, encodes each relation in a compact 16-D token computed only for the points being queried, and still exposes metric displacement through a single linear map that transfers across scenes and datasets. Spatial structure of this kind can thus be acquired with far less supervision and a far more compact representation than reconstruction requires. Whether such relational representations can support reasoning, planning, or control, and whether RePT extends to dynamic environments, are the natural next questions.

Why does this organization emerge? We hypothesize that it reflects the quantity that stays fixed under the training signal. In a static scene, the displacement between two points is invariant across viewpoints while its image-space appearance is not, so a relation that must transfer across views is pushed toward encoding this invariant. Additive relations then compose as displacements do. Our ablations are consistent with this account, as removing distinct-point prediction, adaptive scene context, or additive structure each weakens the organization. The hypothesis predicts that the organization should weaken in non-rigid scenes, where inter-point displacements change across views. Why the resulting structure is linearly accessible remains open.

Our claims concern representation structure rather than reconstruction. Readouts are fitted using ground-truth 3D displacements, and mean endpoint errors remain 0.51--0.90\,m. Default training tracks are constructed offline from depth and pose, although RGB-only tracking yields comparable geometry. Evaluation is limited to static indoor scenes. Finally, the shared readout transfers across scenes only up to a scene-specific rotational gauge, so metric endpoint evaluation still requires estimating one rotation per scene. Whether cross-scene alignment can emerge without this calibration is open.

\section{Conclusion}
We introduced RePT, a framework for learning cross-view spatial relations from 2D observations without explicit 3D prediction targets. Despite having no 3D prediction target, the resulting relation space develops a structured correspondence with underlying 3D geometry. Relation magnitude and direction track metric 3D displacement, local neighborhoods partially reflect 3D proximity, and a shared linear projection maps the learned relations into a metric 3D space up to a scene-specific rotation. Controlled experiments show that distinct-point relational prediction consistently strengthens this geometric organization, which also depends on additive relations and adaptive scene context, and higher point-transfer accuracy does not imply stronger geometry. The organization persists on ScanNet++ and with RGB-only correspondences. Together, these results suggest a complementary route toward geometry-aware visual representations without 3D prediction targets.

\section*{LLM Usage Statement}
Generative AI tools (ChatGPT, Codex, and Claude Code) were used to assist with implementation of the proposed method and experimental pipeline, and interpretation of experimental results. They were also used for manuscript drafting and editing, paper organization, literature search and summarization, and the design and refinement of scientific figures. The core research questions, conceptual framework, and hypotheses were developed by the authors, and the experimental design, including all baselines and controlled comparisons, was conceived and designed by the human authors. All AI-generated code, research suggestions, analyses, and manuscript content were reviewed and verified by the authors. The authors accept \mbox{full responsibility for this paper}.

%% file: appendix.tex
\section{Dataset Construction}
\label{app:exp_details}

We describe the offline landmark-track construction and episode sampling used for RePT. The ScanNet++~\citep{yeshwanth2023scannetpp} evaluation in the main text uses the same preprocessing and sampling procedure as ARKitScenes~\citep{baruch2021arkitscenes}.

\subsection{Landmark Track Construction}
\label{app:track_construction}
We build persistent tracks for each RGB-D sequence.

\paragraph{Initialization.}
In the first frame, we divide the image into a regular grid and randomly sample a valid-depth pixel in each cell. Each sampled pixel $\mathbf{u}=(u,v)$ is back-projected into world coordinates using its depth $d_t(\mathbf{u})$, the camera intrinsics $K_t$, and the camera-to-world pose $(R_t,\mathbf{t}_t)$:
\begin{equation}
    \mathbf{X}_i=R_t\left[d_t(\mathbf{u})K_t^{-1}\tilde{\mathbf{u}}\right]+\mathbf{t}_t,
    \qquad \tilde{\mathbf{u}}=(u,v,1)^\top.
    \label{eq:track_backprojection}
\end{equation}
Each landmark receives a unique identifier $i$ and a fixed world position $\mathbf{X}_i$. Its visibility and 2D location in the current frame are recorded as its initial observation.

\paragraph{Propagation and replenishment.}
We process the remaining frames in temporal order. In frame $s$, the accumulated landmarks are transformed into the current camera coordinates and reprojected onto the image:
\begin{equation}
    \mathbf{x}_{i,s}=R_s^\top(\mathbf{X}_i-\mathbf{t}_s),
    \qquad \mathbf{u}_{i,s}=\pi(K_s\mathbf{x}_{i,s}),
    \label{eq:track_reprojection}
\end{equation}
where $\pi([a,b,c]^\top)=(a/c,b/c)^\top$ denotes perspective division. We first project each landmark into the current image, retaining only points in front of the camera and within the image boundaries. We then verify visibility by comparing each point's projected depth with the observed depth in a small neighborhood around its projected location. Visible landmarks retain their identifiers and provide new 2D observations. We reuse existing points in occupied grid cells and sample new valid-depth pixels in empty cells, creating additional landmarks through \autoref{eq:track_backprojection}. This procedure preserves point identities across views while replenishing coverage as the camera moves.

\paragraph{Track formation.}
Repeating this process yields a set of landmarks with persistent identifiers, per-frame visibility, and 2D image coordinates. Depth and camera parameters are used only during this offline stage to establish cross-view point correspondences.

RePT is trained using images and the resulting 2D point observations.
Depth and camera parameters establish the offline correspondences; they are
not provided to the model or used as prediction targets.

\subsection{RGB-Only Correspondence Construction}
\label{app:rgb_only_tracks}

To test an alternative source of training correspondences, we process
ARKitScenes with CoTracker3~\citep{karaev2025cotracker3}, an RGB-only point
tracker. Instead of sampling valid-depth pixels, we randomly sample query
points from RGB pixels without using depth. CoTracker3 tracks these points
through RGB frames in place of depth-and-pose projection.
We use the online CoTracker3 model with a 16-frame window and an update step
of 8, process every RGB frame, seed new queries in empty cells of a
$12\times8$ grid, and retire a track after 32 consecutive frames without a
visible prediction. Only RGB frames are read.
The resulting 2D trajectories are used with the
same temporal sampling, episode construction, and RePT training procedure.
We fully train a separate RePT model on these tracks.

We evaluate both models on the same 55 ARKitScenes validation scenes,
relation pairs, and 300 random 16/39 scene splits.
\autoref{tab:rgb_only_geometry_full} reports the complete geometric comparison
corresponding to \autoref{tab:rgb_only_geometry}.
CoTracker3 correspondences yield higher raw-token correlations and neighborhood
overlap, while linear readouts retain comparable geometric structure.
For held-out transfer, 3D endpoint error is 0.9130\,m, compared with
0.8962\,m for projection-based construction.
Over the same 300 splits, the held-out transfer differences relative to
projection-based construction are $-0.0075\pm0.0044$ in length correlation,
$-0.0161\pm0.0047$ in angle correlation, and $+0.0168\pm0.0083$\,m in endpoint
error.
These results show that the geometric organization also emerges when training
correspondences are obtained by RGB-only tracking.

Tracking all 5,047 ARKitScenes sequences (3.99M frames), which include our
training and validation sequences, takes 28.8 GPU-hours
on a single NVIDIA H20 GPU, so RePT extends efficiently to video collections
without camera-pose or depth annotations. We use depth-and-pose projection for the main
experiments and controlled comparisons, and report RGB-only tracking as an
alternative correspondence source.

\begin{table*}[t]
    \centering
    \caption{
        \textbf{Complete geometric evaluation with RGB-only training correspondences.}
        Both RePT models use the same training pipeline; only point sampling
        and correspondence construction change.
        Evaluation uses the same scenes, relation pairs, and 300 random 16/39
        scene splits. Len.\ and Ang.\ denote scene-averaged Pearson
        correlations, and NN denotes NN-Overlap@10.
    }
    \label{tab:rgb_only_geometry_full}
    \scriptsize
    \setlength{\tabcolsep}{4pt}
    \begin{tabular*}{\textwidth}{@{\extracolsep{\fill}}lcccccccc@{}}
        \toprule
        & \multicolumn{4}{c}{Depth-and-pose projection}
        & \multicolumn{4}{c}{RGB-only CoTracker3} \\
        \cmidrule(lr){2-5}\cmidrule(lr){6-9}
        Representation / Readout
        & Len. $\uparrow$ & Ang. $\uparrow$ & NN $\uparrow$ & 3D Err. (m) $\downarrow$
        & Len. $\uparrow$ & Ang. $\uparrow$ & NN $\uparrow$ & 3D Err. (m) $\downarrow$ \\
        \midrule
        Raw $q$ & 0.5921 & 0.6789 & 0.2639 & --
                & 0.6276 & 0.7088 & 0.3051 & -- \\
        Scene-specific $W_s$ & 0.8639 & 0.8804 & 0.4657 & 0.5136
                            & 0.8598 & 0.8558 & 0.4379 & 0.5353 \\
        Shared $W$ & 0.7349 & 0.7979 & 0.3721 & 0.8637
                   & 0.7286 & 0.7816 & 0.3532 & 0.8770 \\
        Transfer $W$ & 0.7168 & 0.7844 & 0.3604 & 0.8962
                     & 0.7093 & 0.7683 & 0.3422 & 0.9130 \\
        \bottomrule
    \end{tabular*}
\end{table*}

\subsection{Training Episode Construction}
\label{app:episode_sampling}
Training episodes sample temporal windows and cross-view point observations from the tracks.

\paragraph{Temporal subsequences.}
We divide each recording into ten interleaved subsequences according to the source frame index modulo ten. Each subsequence follows the indices $r,r+10,r+20,\ldots$, with $r\in\{0,\ldots,9\}$, and all ten offsets are used for sampling. For recordings too short to support this subdivision, we use the full valid frame stream instead.

\paragraph{Episode construction.}
Each episode samples a contiguous 13-frame window within a subsequence, using the first 12 images to construct $C$ and reserving the final image as the auxiliary reconstruction target. From the tracks observed in the first 12 frames, we select two distinct landmarks $A$ and $B$ and sample four observation pairs to encode their relation and up to eight additional pairs as point-transfer queries. Each pair observes $A$ and $B$ in different frames. Relation and query sets use distinct observation pairs, while allowing shared images.

\section{Geometric Evaluation Protocol}
\label{app:linear_readout}

We evaluate the frozen relation representation using raw tokens,
scene-specific linear readouts, a shared linear readout, and transfer of that
shared readout to held-out scenes. This section specifies the relation
sampling, readout fitting, and evaluation metrics used in the main text.

\subsection{Scene-Specific Readout}
\label{app:scene_specific_readout}

For each scene, we sample 1,500 valid cross-view relation pairs independently
from arbitrary valid-depth image locations, rather than from the landmark
tracks used to construct training episodes. Each relation token is encoded
from a single observation pair, without the averaging over four observation
pairs used during training. We denote each relation token by $q_i\in\mathbb{R}^{16}$ and its 3D
displacement by \mbox{$\mathbf d_i=\mathbf x_{B_i}-\mathbf x_{A_i}\in\mathbb{R}^{3}$}.

For a scene-specific readout, we randomly divide the relations into 1,200
fitting pairs and 300 test pairs. We fit a bias-free linear map
$W_s\in\mathbb{R}^{16\times3}$ by least squares:
\begin{equation}
W_s=\arg\min_W\left\|Q_s^{\mathrm{fit}}W-D_s^{\mathrm{fit}}\right\|_F^2.
\end{equation}
Here, $Q_s^{\mathrm{fit}}$ and $D_s^{\mathrm{fit}}$ stack the fitting
relation tokens and their displacement targets, respectively.
The fitted map remains fixed when predicting
$\hat{\mathbf d}_i=q_iW_s$ on the test pairs. Raw-token metrics use the
unprojected relation representation and require no fitted readout.

\begin{table*}[t]
\centering
\caption{\textbf{Variability over 300 random 16/39 scene splits.}
Mean $\pm$ standard deviation. Len.\ and Ang.\ are scene-averaged Pearson
correlations, and NN denotes NN-Overlap@10. The zero-displacement row predicts
$\hat{\mathbf d}=\mathbf 0$.}
\label{tab:variability}
\scriptsize
\setlength{\tabcolsep}{4pt}
\begin{tabular*}{\textwidth}{@{\extracolsep{\fill}}lccccc@{}}
\toprule
Representation / Readout & Len. $\uparrow$ & Ang. $\uparrow$ & NN $\uparrow$ & 3D Err. (m) $\downarrow$ & $R^2$ $\uparrow$ \\
\midrule
Zero displacement     & -- & -- & -- & 1.7324 & -- \\
Raw $q$               & $0.5921\pm0.0219$ & $0.6789\pm0.0131$ & $0.2639\pm0.0069$ & -- & -- \\
Scene-specific $W_s$  & $0.8639\pm0.0171$ & $0.8804\pm0.0133$ & $0.4657\pm0.0192$ & $0.5136\pm0.0446$ & $0.8931\pm0.0136$ \\
Shared $W$            & $0.7349\pm0.0274$ & $0.7979\pm0.0181$ & $0.3721\pm0.0171$ & $0.8637\pm0.0710$ & $0.6919\pm0.0637$ \\
Transfer $W$          & $0.7168\pm0.0140$ & $0.7844\pm0.0071$ & $0.3604\pm0.0070$ & $0.8962\pm0.0203$ & $0.6653\pm0.0367$ \\
\bottomrule
\end{tabular*}
\end{table*}

\subsection{Shared Readout and Held-Out Transfer}

For scene $s$, let
$Q_s\in\mathbb{R}^{N_s\times16}$ contain the relation tokens and
$D_s\in\mathbb{R}^{N_s\times3}$ their corresponding physical displacements
$\mathbf{x}_B-\mathbf{x}_A$. We jointly fit a bias-free shared linear map
$W\in\mathbb{R}^{16\times3}$ and per-scene rotations
$R_s\in\mathrm{SO}(3)$:
\begin{equation}
\min_{W,\{R_s\}}
\sum_s
\left\|
Q_s W R_s^\top-D_s
\right\|_F^2.
\label{eq:shared_readout}
\end{equation}
Neither the relation tokens nor the displacement targets are centered, so the
readout is origin-preserving. The scene-specific rotations account only for
differences in the orientation of the independently defined scene coordinate
frames; no translation, scale, or reflection is fitted.

\paragraph{Alternating optimization.}
We optimize \autoref{eq:shared_readout} by alternating between $W$ and $\{R_s\}$. All rotations are initialized as $R_s=I$. With the rotations fixed,
\[
Q_s W R_s^\top \approx D_s
\quad\Longleftrightarrow\quad
Q_s W \approx D_s R_s,
\]
so stacking all scenes updates $W$ by least squares:
\begin{equation}
W
=
\arg\min_W
\sum_s
\left\|
Q_sW-D_sR_s
\right\|_F^2.
\end{equation}
With $W$ fixed, we compute $\hat D_s=Q_sW$ and independently update each rotation using the Kabsch algorithm:
\begin{equation}
R_s
=
\arg\min_{R\in\mathrm{SO}(3)}
\left\|
\hat D_sR^\top-D_s
\right\|_F^2.
\end{equation}
The alignment allows only proper 3D rotations.

Convergence is determined using only the fitting split, and the held-out 20\% test pairs are never used during optimization. After each alternating update, we compute
\begin{equation}
L_t
=
\sum_s
\left\|
Q_sW_tR_{s,t}^{\top}-D_s
\right\|_F^2,
\end{equation}
and stop when
\[
|L_t-L_{t-1}|
\le
10^{-7}\max(|L_{t-1}|,10^{-12}),
\]
or after 50 iterations. The final prediction is
\begin{equation}
\hat{\mathbf d}_{A\rightarrow B}
=
q_{A\rightarrow B} W R_s^\top.
\end{equation}
Full RePT converges in about 7 alternating steps.

\paragraph{Shared-readout evaluation.}
Each split randomly partitions the 55 scenes into 16 readout-fitting scenes
and 39 held-out scenes. We use the same relation sampling and
\mbox{1,200/300} fitting/test partition.

For each of the 16 fitting scenes, we use 1,200 relation pairs to jointly fit
the shared linear map $W$ and the scene-specific rotation $R_s$, yielding
19,200 fitting pairs in total. The remaining 300 pairs per scene are held out
for evaluation. Both $W$ and $R_s$ are fixed when evaluating these held-out
pairs. Results are averaged over 300 random scene splits (16 fitting, 39 held out).

\paragraph{Held-out transfer evaluation.}
For each split, the shared $W$ fitted from the 16 readout-fitting scenes is
frozen and applied unchanged to the remaining 39 held-out scenes. For metrics
that are invariant to scene rotation, such as length and angle correlation, no
additional alignment is required.

For 3D endpoint error, we estimate only the scene rotation $R_s$ by Kabsch
alignment on 1,200 relation pairs per held-out scene, keeping $W$ fixed.
The predictions for the 300 test pairs are:
\begin{equation}
\hat{\mathbf d}_{s}
=
q_{s} W R_{s}^{\top}.
\end{equation}
The shared map therefore remains unchanged on held-out scenes, while
$R_s$ accounts for the orientation of each scene's coordinate frame.

\subsection{Geometric Metrics}

\paragraph{Length and angle correlations.}
For raw tokens, we compare $\|q_i\|_2$ with $\|\mathbf d_i\|_2$, and
angles between relation tokens with angles between their corresponding
physical displacement vectors. For linear readouts, we make the same
comparisons using the predicted 3D displacements in place of $q_i$.
We report Pearson correlations for linear association and Spearman
correlations for rank agreement. Correlations are computed within each scene
and averaged across scenes. Shared and held-out transfer results are further
averaged over the 300 random 16/39 scene splits.

\paragraph{3D endpoint error.}
We report the mean Euclidean distance between predicted and ground-truth
3D displacement vectors, measured in meters:
\begin{equation}
\operatorname{Err}_{\mathrm{3D}}
=\frac{1}{N}\sum_{i=1}^{N}
\left\|\hat{\mathbf d}_i-\mathbf d_i\right\|_2.
\end{equation}
For shared and transfer readouts, $\hat{\mathbf d}_i=q_iWR_s^\top$
uses the scene rotation estimated from fitting pairs. Length and angle
correlations are invariant to this rotation. We do not evaluate 3D endpoint error on raw relation tokens.

\paragraph{NN-Overlap@$k$.}
NN-Overlap measures whether local neighborhoods in a relation representation
match those in physical 3D displacement space. For relation $i$, let
$\mathcal N_k^{r}(i)$ be its $k$ nearest neighbors under Euclidean distance in
the evaluated representation (raw $q$ or a linear readout), and let
$\mathcal N_k^{3D}(i)$ be the $k$ nearest neighbors of its ground-truth 3D
displacement. We compute
\begin{equation}
    \operatorname{NN\text{-}Overlap}@k
    = \frac{1}{N}\sum_{i=1}^{N}
    \frac{\left|\mathcal N_k^{r}(i)\cap
    \mathcal N_k^{3D}(i)\right|}{k}.
\end{equation}
We use $k=10$ and compute neighborhoods within each scene. A higher score
indicates better preservation of local 3D neighborhood structure.

\begin{figure*}[t]
\centering
\includegraphics[width=\linewidth]{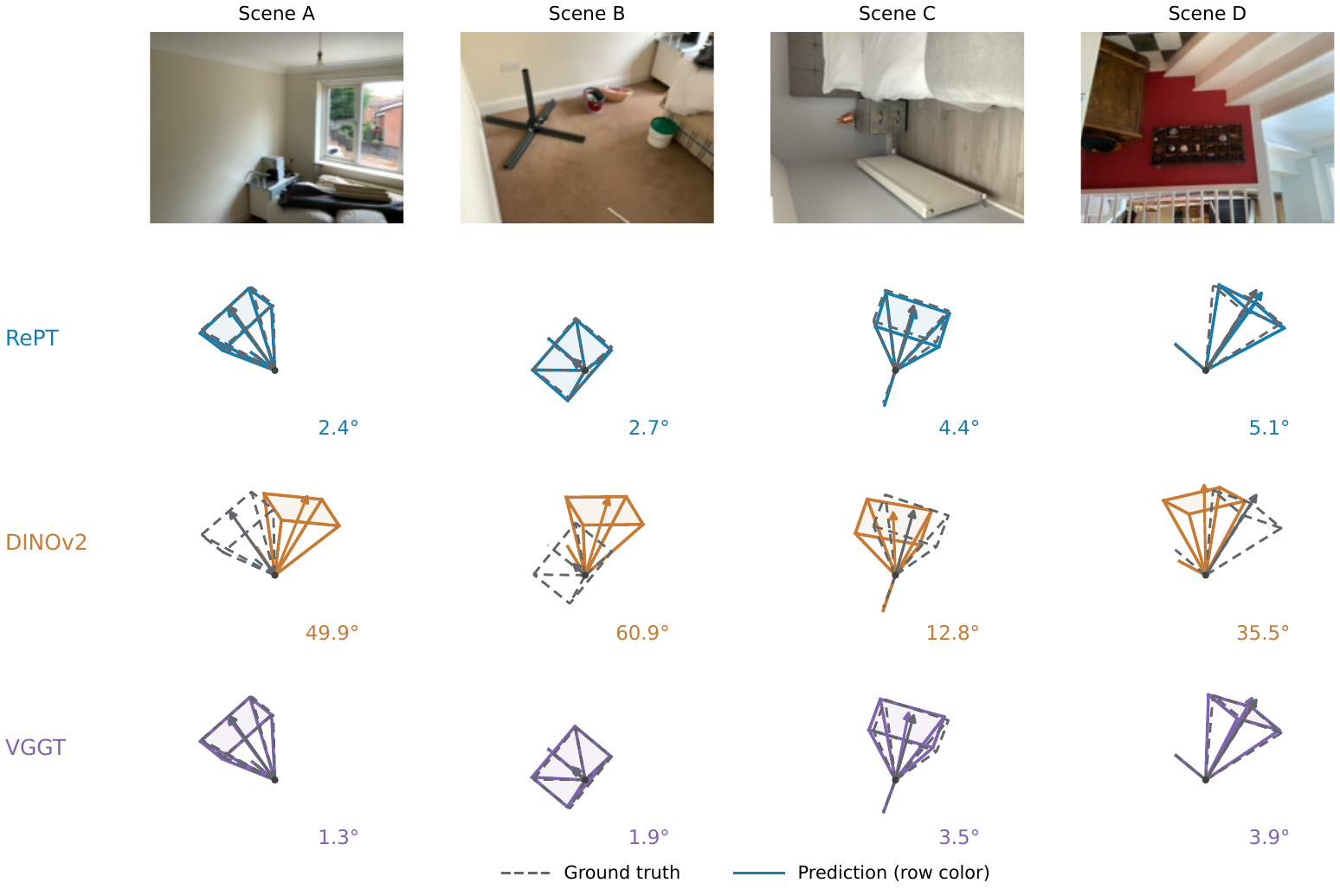}
\caption{\textbf{Camera orientation on four ARKitScenes scenes.} Dashed gray cameras show ground truth; colored cameras show predictions relative to the first frame. Camera centers coincide to isolate orientation. Each column is a different validation scene and shows its RGB frame; rows compare RePT, DINOv2, and VGGT. Each panel reports its rotation error.}
\label{fig:ark_camera_orientation}
\end{figure*}

\subsection{Point-Transfer Metrics}

Here, we describe the point-transfer metrics used in the main text. Let
$\hat{\mathbf p}_i$ and $\mathbf p_i^*$ denote the predicted and ground-truth
2D locations for query $i$, respectively, and let $N$ be the number of queries.

\paragraph{End-Point Error (EPE).}
EPE is the Euclidean distance between a predicted point and its ground-truth
location, measured in pixels of the original $192\times256$ ARKitScenes frames. We report its
mean over all queries:
\begin{equation}
    \operatorname{EPE}
    = \frac{1}{N}\sum_{i=1}^{N}
    \left\lVert \hat{\mathbf p}_i-\mathbf p_i^* \right\rVert_2.
\end{equation}
Lower EPE indicates more accurate point transfer.

\paragraph{PCK@$\tau$.}
Percentage of Correct Keypoints (PCK) measures the fraction of predictions
whose EPE is no greater than a pixel threshold $\tau$:
\begin{equation}
    \operatorname{PCK}@\tau
    = \frac{1}{N}\sum_{i=1}^{N}
    \mathbf{1}\!\left[
    \left\lVert \hat{\mathbf p}_i-\mathbf p_i^* \right\rVert_2
    \leq \tau\right].
\end{equation}
Higher values indicate that a larger fraction of predictions fall within the
specified tolerance. The main text reports PCK@4; \autoref{tab:point_transfer_full}
additionally reports EPE, PCK@8, and PCK@16.

\autoref{tab:point_transfer_full} reports these metrics for the variants in \autoref{tab:controlled_ablation}, on the 2,164 ARKitScenes
validation episodes. For the correspondence baseline, the learned endpoint
embeddings are evaluated under the same $A\rightarrow B$ relational transfer
protocol as RePT: the relation is formed from four $(A,B)$ observation pairs
and the trained decoder predicts $B$ in a new target view.
The MLP-relation variant achieves the best point transfer on every metric
while showing the weakest geometric organization among the variants with
adaptive context.

\begin{table}[t]
\centering
\caption{\textbf{Point-transfer metrics for the controlled variants.}
EPE is in pixels of the original $192\times256$ frames.}
\label{tab:point_transfer_full}
\scriptsize
\setlength{\tabcolsep}{3pt}
\begin{tabular*}{\columnwidth}{@{\extracolsep{\fill}}lccccc@{}}
\toprule
& \multicolumn{2}{c}{EPE (px) $\downarrow$} & \multicolumn{3}{c}{PCK $\uparrow$} \\
\cmidrule(lr){2-3}\cmidrule(lr){4-6}
Method & Mean & Median & @4 & @8 & @16 \\
\midrule
Full RePT        & 8.87  & 6.41  & 0.2789 & 0.6136 & 0.8741 \\
Correspondence   & 12.94 & 8.92  & 0.2034 & 0.4540 & 0.7430 \\
DINOv2 as $C$    & 20.24 & 15.48 & 0.0567 & 0.1919 & 0.5183 \\
Frozen context   & 33.84 & 21.62 & 0.0667 & 0.1889 & 0.3927 \\
MLP relation     & 8.28  & 5.91  & 0.3083 & 0.6493 & 0.8886 \\
\midrule
\multicolumn{6}{@{}l}{\emph{Training from scratch}} \\
Relation + recon. & 8.47 & 6.03 & 0.3055 & 0.6391 & 0.8827 \\
Relation only     & 9.11 & 6.53 & 0.2742 & 0.5944 & 0.8659 \\
\bottomrule
\end{tabular*}
\end{table}

\subsection{Variability and Reference Baselines}
\label{app:variability}

\autoref{tab:variability} reports the mean and standard deviation of the
main-text metrics over the 300 random 16/39 scene splits, together with the
coefficient of determination
$R^2=1-\sum_i\|\hat{\mathbf d}_i-\mathbf d_i\|_2^2/\sum_i\|\mathbf d_i-\bar{\mathbf d}\|_2^2$,
computed within each scene on test pairs and averaged across scenes.
Ground-truth displacements have a mean length of 1.7324\,m (median 1.59\,m),
which is also the endpoint error of predicting a zero displacement.
The scene-specific readout reaches an $R^2$ of 0.8931, and the shared readout
transferred to held-out scenes reaches 0.6653; its endpoint error of 0.8962\,m
is roughly half that of the zero-displacement prediction.

\section{Geometry in the Context Representation}
\label{app:c_geometry}

We probe the frozen scene representation $C$ for depth, camera pose, and spatial relations to complement the relation-space analysis. We compare against the frozen DINOv2~\citep{oquab2023dinov2} features used as input to the Context Encoder, and against VGGT~\citep{wang2025vggt}, a reference representation trained with explicit geometric objectives; see \autoref{tab:context_geometry} for results.

\subsection{Depth and Relative Camera Pose}
We freeze RePT, DINOv2, and VGGT and train matched shallow transformer probes on ARKitScenes to predict depth and camera pose relative to the first frame. Each representation is mapped to 256 channels by a linear adapter, followed by an identical two-block transformer probe (width 256, 4 heads); the probes receive only the frozen visual features. Probes are trained on one 12-frame clip from each of the 4,492 training sequences for 30 epochs and evaluated on four clips from each of the 55 validation scenes. A constant-feature probe provides a lower reference. Depth is evaluated without scale alignment. We report depth RMSE in meters and median angular errors for camera rotation and translation direction. \autoref{fig:ark_camera_orientation} compares camera orientations predicted by the frozen geometric probes across four distinct ARKitScenes validation scenes.

\subsection{Point-Level Spatial Relations}
We additionally probe local spatial relations following SpaRRTa~\citep{kargin2026sparrta}. We adapt the task to tracked point pairs and train an MLP to classify whether $B$ lies in front of, to the right of, behind, or to the left of $A$ in the camera frame of a reference view. The classifier receives the two endpoint features together with a frame descriptor obtained by mean-pooling the features of the reference view. In the egocentric (ego) setting, both endpoints are observed in the reference view. In the allocentric (allo) setting, $A$ and $B$ are observed in other views and classified in a distinct reference view. We report accuracy on held-out scenes; chance is 25\%.

\begin{table}[t]
\centering
\caption{\textbf{Geometric information in frozen features on ARKitScenes.}
Pose errors are in degrees and depth RMSE is in meters. Spatial accuracy
is averaged over three probe seeds (standard deviation at most 0.008). DINOv2 and VGGT
use official pretrained checkpoints.}
\label{tab:context_geometry}
\scriptsize
\setlength{\tabcolsep}{3pt}
\begin{tabular*}{\columnwidth}{@{\extracolsep{\fill}}lccccc@{}}
\toprule
& \multicolumn{2}{c}{Pose error $\downarrow$} & Depth & \multicolumn{2}{c}{Spatial Acc. $\uparrow$} \\
\cmidrule(lr){2-3}\cmidrule(lr){5-6}
Representation & Rot. & Trans. dir. & RMSE $\downarrow$ & Ego & Allo \\
\midrule
Constant & 51.120 & 79.560 & 0.8509 & -- & -- \\
DINOv2 & 45.640 & 62.843 & 0.4448 & 0.7902 & 0.5320 \\
RePT ($C$) & 13.225 & 32.650 & 0.4380 & 0.8490 & 0.7589 \\
VGGT & 2.355 & 6.098 & 0.2902 & 0.8979 & 0.8783 \\
\bottomrule
\end{tabular*}
\end{table}
As shown in \autoref{tab:context_geometry}, $C$ improves over DINOv2 on all
reported tasks. The depth gain is modest, while pose and spatial classification
improve substantially; the largest gain is in the allocentric setting
(0.7589 vs.\ 0.5320), where the endpoints are observed outside the reference
frame. VGGT achieves the best results across all tasks.

\section{Relation Token Dimensionality}
\label{app:relation_dimensionality}

\autoref{tab:relation_dimensionality} reports geometric organization for relation
tokens of 3, 4, 8, 16, and 32 dimensions under the evaluation protocol of the main text.
Scene-specific length correlation increases with dimension, and 32 dimensions gives the
strongest scene-specific results. Cross-scene readouts do not follow this trend: 3 dimensions
gives the strongest shared and held-out correlations. Among 8, 16, and 32 dimensions, 16 gives the
strongest held-out transfer, with length and angle correlations of 0.7168 and
0.7844 and the lowest transfer endpoint error (0.8962\,m, compared with
0.9078\,m for 8 and 0.9177\,m for 32 dimensions).

\begin{table}[t]
    \centering
    \caption{
        \textbf{Geometric organization across relation token dimensions.}
        Results are averaged over 300 random 16/39 scene splits.
        Len.\ and Ang.\ denote scene-averaged Pearson correlations.
    }
    \label{tab:relation_dimensionality}
    \scriptsize
    \setlength{\tabcolsep}{2pt}
    \begin{tabular*}{\columnwidth}{@{\extracolsep{\fill}}lcccccc@{}}
        \toprule
        & \multicolumn{2}{c}{Scene-specific}
        & \multicolumn{2}{c}{Shared}
        & \multicolumn{2}{c}{Transfer} \\
        \cmidrule(lr){2-3}\cmidrule(lr){4-5}\cmidrule(lr){6-7}
        Dim. & Len. $\uparrow$ & Ang. $\uparrow$ & Len. $\uparrow$ & Ang. $\uparrow$ & Len. $\uparrow$ & Ang. $\uparrow$ \\
        \midrule
        3 & 0.7878 & 0.8468 & \textbf{0.7586} & \textbf{0.8174} & \textbf{0.7557} & \textbf{0.8154} \\
        4 & 0.7902 & 0.8128 & 0.7366 & 0.7773 & 0.7302 & 0.7746 \\
        8 & 0.7911 & 0.8347 & 0.6932 & 0.7801 & 0.6793 & 0.7720 \\
        16 & 0.8639 & 0.8804 & 0.7349 & 0.7979 & 0.7168 & 0.7844 \\
        32 & \textbf{0.8992} & \textbf{0.9039} & 0.7328 & 0.7866 & 0.7044 & 0.7647 \\
        \bottomrule
    \end{tabular*}
\end{table}

\begin{figure*}[t]
    \centering
    \includegraphics[width=\textwidth]
    {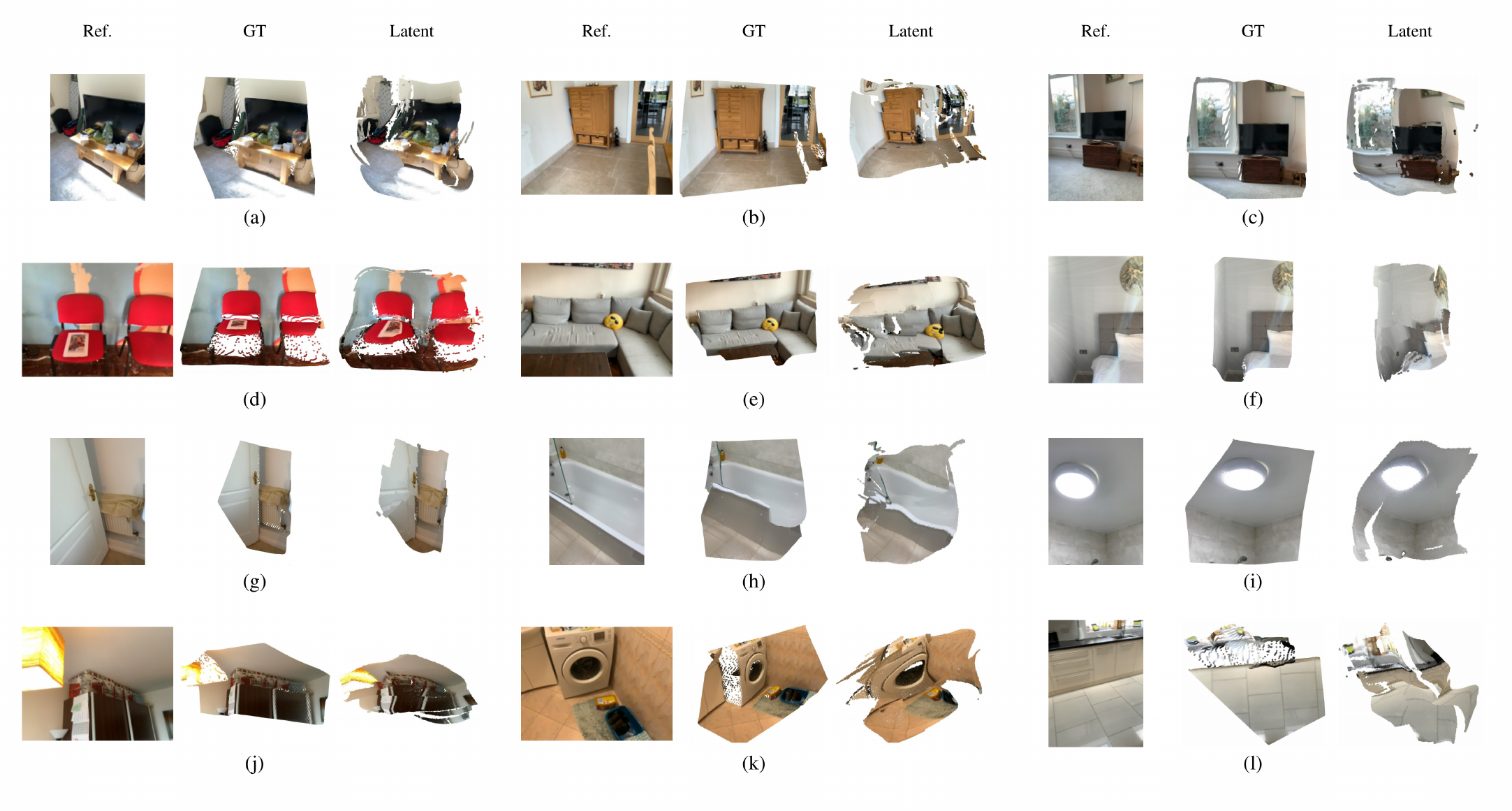}
    \caption{
        \textbf{Additional visualizations of linearly projected geometry.}
        Each example shows a reference image (\emph{Ref.}), the ground-truth
        3D geometry (\emph{GT}), and the geometry recovered by linearly
        projecting the learned relation representation into 3D
        (\emph{Latent}). The bottom row shows examples with more pronounced
        geometric distortions.
    }
    \label{fig:more_geometry_visualizations}
\end{figure*}

\section{Additional Geometry Visualizations}
\label{sec:more_geometry_visualizations}

We visualize relative 3D geometry using the scene-specific linear readout
from Appendix~\ref{app:scene_specific_readout}. For each example, we select a
reference point $A$ and target points $\{P_k\}_{k=1}^{K}$ in an evaluation
scene. We independently encode each directed relation $q_{A\rightarrow P_k}$
from a cross-view observation pair and project it with the fitted $W_s$.
Placing $A$ at the origin gives the predicted and ground-truth relative
positions:
\begin{equation}
\begin{aligned}
\hat{\mathbf x}_{P_k} &= q_{A\rightarrow P_k}W_s, \\
\mathbf x_{P_k}^{\mathrm{GT}} &= \mathbf x_{P_k}-\mathbf x_A.
\end{aligned}
\end{equation}
We sample one target point per pixel at the $168\times224$ input resolution,
yielding up to $168\times224$ relations from the common reference point.
Predicted and ground-truth visualizations use the same anchor and targets;
they differ only in how the relative displacements are obtained.

We repeat this construction independently for each example in
\autoref{fig:latent_geometry_vis} and
\autoref{fig:more_geometry_visualizations}. The decoded point clouds recover
much of the dominant scene structure. Harder cases exhibit local bending,
missing regions, or larger geometric distortions, as illustrated in the
bottom row. These visualizations show relative layouts around the reference
point rather than absolute coordinates.

\section{Implementation Details}
\label{app:implementation}

\autoref{tab:implementation} lists the architecture and optimization settings
of the main RePT model. All variants in \autoref{tab:controlled_ablation} use the same
settings unless stated otherwise. The two from-scratch variants are trained for
15 epochs, matching the total budget of 5 reconstruction-only epochs plus 10
joint epochs used by Full RePT.

\begin{table}[t]
\centering
\caption{\textbf{Architecture and training settings of RePT.}}
\label{tab:implementation}
\scriptsize
\setlength{\tabcolsep}{2pt}
\begin{tabular*}{\columnwidth}{@{\extracolsep{\fill}}ll@{}}
\toprule
Image backbone & DINOv2 ViT-S/14, frozen; $168\times224$ input \\
Context Encoder & 24 blocks (12 within-/inter-view pairs), width 512, 8 heads \\
Scene tokens & 48 per frame ($6\times8$), 12 context frames \\
Relational Enc.\ / Dec. & 4 blocks each, width 512, 8 heads \\
Relation token & 16 dimensions \\
Reconstruction decoder & 4 blocks, width 512; 90\% masking \\
Trainable parameters & 129.5M (frozen backbone 22.1M) \\
\midrule
Optimizer & AdamW, $\beta=(0.9,0.95)$, weight decay 0.1 \\
Learning rate & $2\times10^{-4}$; $5\times10^{-5}$ for $E_{\mathrm{ctx}}$ and $D_{\mathrm{rec}}$ \\
Schedule & 1\% linear warmup, cosine decay \\
Loss & Smooth L1 ($\beta=0.02$), $\lambda_{\mathrm{rec}}=0.1$ \\
Batch size & 128 episodes \\
Training & 10 epochs, 55,630 updates \\
Initialization & 5 epochs of reconstruction-only pretraining \\
Gradient clipping & 1.0 \\
Precision / hardware & bfloat16, 4 NVIDIA H20 GPUs \\
\bottomrule
\end{tabular*}
\end{table}